\documentclass{bmvc2k}

\usepackage[misc]{ifsym}
\newcommand{\corr}{(\Letter)}
\usepackage{xcolor}
\usepackage{hyperref}

\makeatletter

\def\bmv@unhbox#1{%
  \expandafter\unhcopy\csname bmv@box #1 \endcsname
}

\newcommand{\makeappendixtitle}{%
  \def\@title{%
    \scalebox{0.90}[1]{GP-VM$\!\times\!$SMA:}%
    Benchmarking General-Purpose Vision Models and Specialized Architectures for 2D Medical Image Segmentation\\
    --- Supplementary Material ---
  }%
  \bmv@maketitle
  \pdfbookmark[0]{Supplementary Material}{appendix}%
}

\makeatother

\usepackage{orcidlink}
\DeclareRobustCommand{\myorcid}[1]{%
  \begingroup
    \hypersetup{hidelinks}%
    \orcidlink{#1}%
  \endgroup
}

\title{\scalebox{0.90}[1]{GP-VM$\!\times\!$SMA:}\ Benchmarking General-Purpose Vision Models and Specialized Architectures for 2D Medical Image Segmentation}

\addauthor{Vanessa Borst \myorcid{0009-0004-7123-7934} \corr}{vanessa.borst@uni-wuerzburg.de}{1}
\addauthor{Anna Riedmann \myorcid{0000-0002-1969-8563}}{anna.riedmann@uni-wuerzburg.de}{1}
\addauthor{Samuel Kounev \myorcid{0000-0001-9742-2063}}{samuel.kounev@uni-wuerzburg.de}{1}

\addinstitution{
 Institute of Computer Science\\
 Julius-Maximilians-University Würzburg\\
 97070 Würzburg, Germany
}

\runninghead{Borst et al.}{Benchmarking GP-VMs and SMAs for 2D MIS}

\usepackage{acro}
\usepackage[inline]{enumitem}
\usepackage[flushleft]{threeparttable} 
\usepackage{threeparttablex}   
\usepackage{multirow}
\usepackage{graphicx}
\usepackage{adjustbox}
\usepackage{booktabs}
\usepackage{rotating}
\usepackage{hyperref}
\usepackage{numprint}
\usepackage{amssymb}
\usepackage{todonotes}
\usepackage{mfirstuc}

\acsetup{list/display=used}  
\DeclareAcronym{msb}{
  short = \textit{MSB},
  long  = \textit{MedSegBenchmarker}
}
\DeclareAcronym{unet}{
  short = U-Net,
  long  = U-Net
}
\DeclareAcronym{hiformer}{
  short = HiFormer-B,
  long  = HiFormer-B
}
\DeclareAcronym{hiformerRN}{
  short = HiFormer-B$^*$,
  long  = HiFormer-B-RN101
}
\DeclareAcronym{missformer}{
  short = MISSFormer,
  long  = MISSFormer
}
\DeclareAcronym{segformer}{
  short = SegFormer,
  long  = SegFormer-B3
}
\DeclareAcronym{segnext}{
  short = SegNeXt,
  long  = SegNeXt-L
}
\DeclareAcronym{swinumamba}{
  short = SU-Mamba,
  long  = Swin-UMamba
}
\DeclareAcronym{vwmit}{
  short = VW-MiT,
  long  = VWFormer+MiT-B3
}
\DeclareAcronym{vwconv}{
  short = VW-Conv,
  long  = VWFormer+ConvNeXt-S
}
\DeclareAcronym{vwformer}{
  short = VWFormer,
  long  = VWFormer
}
\DeclareAcronym{internimage}{
  short = InternImage,
  long  = InternImage-T
}
\DeclareAcronym{internimageUperhead}{
  short = InternImage,
  long  = InternImage-T+UPerHead
}
\DeclareAcronym{transnext}{
  short = TransNeXt,
  long  = TransNeXt-Tiny
}
\DeclareAcronym{transnextUperhead}{
  short = TransNeXt,
  long  = TransNeXt-Tiny+UPerHead
}
\DeclareAcronym{uperhead}{
  short = UPerHead,
  long  = UPerHead
}
\DeclareAcronym{ukan}{
  short = U-KAN,
  long  = U-KAN-L
}
\DeclareAcronym{cenet}{
  short = CENet$^*$,
  long  = CENet with PVTv2-B3
}
\DeclareAcronym{sota}{
  short = SotA,
  long  = state-of-the-art
}
\DeclareAcronym{ood}{
  short = OOD,
  long  = out-of-distribution
}
\DeclareAcronym{gp}{
  short = GP,
  long  = general-purpose
}
\DeclareAcronym{cnn}{
  short = CNN,
  long  = convolutional neural network
}
\DeclareAcronym{vit}{
  short = ViT,
  long  = vision transformer
}
\DeclareAcronym{gmac}{
  short = GMAC,
  long  = Giga Multiply-Accumulate Operation
}
\DeclareAcronym{gradcam}{
  short = Grad-CAM,
  long  = Gradient-weighted Class Activation Mapping
}
\DeclareAcronym{hcp}{
  short = HCP,
  long  = healthcare professional
}
\DeclareAcronym{cv}{
  short = CV,
  long  = cross-validation
}
\DeclareAcronym{dl}{
  short = DL,
  long  = deep learning
}
\DeclareAcronym{gt}{
  short = GT,
  long  = ground truth
}
\DeclareAcronym{bkai}{
  short = NeoPolyp,
  long  = BKAI-IGH NeoPolyp Small
}
\DeclareAcronym{isic}{
  short = ISIC'18,
  long  = ISIC'18
}
\DeclareAcronym{camus}{
  short = CAMUS,
  long  = CAMUS 
}
\DeclareAcronym{mis}{
  short = MIS,
  long  = medical image segmentation
}
\DeclareAcronym{ss}{
  short = SS,
  long  = semantic segmentation
}
\DeclareAcronym{xai}{
  short = XAI,
  long  = explainability
}
\DeclareAcronym{gpvm}{
  short = GP-VM,
  long  = general-purpose vision model
}
\DeclareAcronym{sma}{
  short = SMA,
  long  = specialized medical segmentation architecture
}
\DeclareAcronym{kan}{
  short = KAN,
  long  = Kolmogorov–Arnold Network
}
\DeclareAcronym{ci}{
  short = CI,
  long  = confidence interval
}
\DeclareAcronym{mlp}{
  short = MLP,
  long  = multilayer perceptron
}

\newif\ifanonymous
\anonymousfalse      
\ifanonymous
  \newcommand{\githuburl}{https://anonymous.4open.science/r/GPVision4MIS-837B}
\else
  \newcommand{\githuburl}{https://github.com/VanessaBorst/GPVMxSMA}
\fi

\begin{document}

\maketitle
\begin{abstract}
\Ac{mis} is a fundamental component of computer-assisted diagnosis and clinical decision support. Over the past decade, numerous architectures specifically tailored to medical imaging have emerged to address domain-specific challenges such as low contrast, small anatomical structures, and limited annotated data. 
In parallel, rapid progress in computer vision has produced highly capable \acp{gpvm} originally designed for natural images. Despite their strong performance on standard vision benchmarks, their effectiveness for \ac{mis} remains insufficiently understood. 
In this controlled empirical study, we examine whether \acp{sma} provide systematic advantages over modern \acp{gpvm} for 2D \ac{mis}. We compare eleven \acp{sma} and \acp{gpvm} using a unified training and evaluation protocol across three heterogeneous datasets, each covering different 2D imaging modalities, class structures, and data characteristics. 
Beyond segmentation performance, we employ linear mixed-effects models (LMMs) for statistical assessment and qualitative \acs{gradcam} visualizations to investigate \ac{xai}-related model behavior. 
Our results imply that, for the analyzed settings, \acp{gpvm} achieve performance comparable to that of specialized MIS models. Moreover, \ac{xai} analyses indicate that \acp{gpvm} are capable of identifying clinically relevant structures despite the absence of explicit domain-specific architectural priors.
These findings suggest that \acp{gpvm} can represent a viable alternative to domain-specific methods for 2D \ac{mis}, highlighting the importance of informed model selection. All code and resources are available at \href{\githuburl}{GitHub}.
\end{abstract}

\section{Introduction}
\label{sec:intro}
Accurate segmentation of medical images plays a central role in computer-assisted diagnosis, treatment planning, and longitudinal disease monitoring. The emergence of deep learning has fundamentally transformed this field, enabling increasingly precise delineation of anatomical structures and pathological regions across imaging modalities. 
Since the introduction of the seminal U-Net architecture~\cite{ronneberger2015unet}, the medical imaging community has produced a variety of domain-specific architectures~\cite{siddique2021u}. These approaches aim to address challenges inherent to medical data, including the detection of small target structures, limited availability of annotated datasets, severe class imbalance, and robustness under challenging clinical conditions characterized by variable image quality and domain-specific artifacts. Many proposed methods build upon the U-Net paradigm while incorporating advances from modern deep learning research, including 
transformer-based mechanisms~\cite{huang2022missformer}, multi-scale feature modeling~\cite{heidari2023hiformer}, state-space layers for long-range dependency modeling~\cite{liuSwinUMambaMICCAI2024}, and alternative architectural paradigms such as \acp{kan}~\cite{li2025ukan}.

In parallel, the broader computer vision community has witnessed rapid progress in powerful \acp{gpvm} designed for dense prediction tasks, such as \ac{ss} and object detection. 
Modern backbones~\cite{shi2024transnext,wang2023internimage} leverage large-scale pretraining on natural image corpora followed by lightweight fine-tuning for downstream tasks. Such models, potentially combined with advanced decoders~\cite{yan2024vwformer}, demonstrate strong performance on challenging benchmarks such as ADE20K~\cite{zhou2017ade20k}, usually exhibiting robust generalization. 
Notably, \acp{gpvm} often benefit from extensive optimization, comprehensive ablation studies, and validation on millions of images---resources that are rarely available for \ac{mis}. 

This progress raises an open question in \ac{mis} regarding the necessity of domain-specific architectural design, as opposed to leveraging increasingly capable \acp{gpvm}.
Initial studies have begun exploring transfer learning and cross-domain adaptation in medical imaging. For example, the Segment Anything Model (SAM)~\cite{kirillov2023SAM}, a milestone in \acs{gp} \ac{ss}, has been adapted to medical contexts~\cite{zhang2024segment,ali2025review}. Other work has demonstrated that large-scale ImageNet-pretraining can yield robust feature representations for medical analysis~\cite{liuSwinUMambaMICCAI2024}. Recent research in wound segmentation has reported that \acp{gpvm} can compete with medical approaches and even challenge-winning wound-targeted methods~\cite{borst2025woundambit}. 
Concurrently, surveys have begun to explore the broader role of generalist models in healthcare. Some studies focus specifically on SAM and its successors~\cite{ali2025review}, whereas others examine generalist approaches in healthcare more broadly~\cite{khan2025comprehensive}. Notably, a 2026 review systematically evaluated generalist models for \ac{mis}, comparing them against task-specific architectures across multiple anatomical targets, and reported that generalist models frequently achieve top-tier performance~\cite{moglia2026generalistModelsInMIS}.

Despite these encouraging findings, certain limitations remain. Survey studies typically rely on metrics reported in original publications~\cite{moglia2026generalistModelsInMIS}, which must be interpreted cautiously due to the absence of standardized evaluation protocols. In practice, studies often differ substantially in datasets, preprocessing pipelines, augmentation strategies, optimization settings, and evaluation procedures, all of which can strongly influence reported performance~\cite{renard2020variability,varoquaux2022machine}. Consequently, performance gains attributed to architectural innovations may instead arise from experimental design choices rather than intrinsic model superiority. Indeed, prior work suggests that newly proposed methods are often not superior to existing implementations once more extensive hyperparameter tuning is performed or alternative random initializations are considered, as summarized by Pineau et al.~\cite{pineau2021improving}.

Extending this concern, the question of whether reported improvements truly reflect model superiority is further complicated by limitations in statistical reporting practices. 
Many segmentation studies report only mean or median performance scores, drawing conclusions directly from these point estimates while neglecting performance variability and statistical uncertainty. In the \ac{mis} community, a comprehensive analysis of MICCAI 2023 segmentation papers showed that over 50\% of studies did not report any assessment of variability, and fewer than 1\% provided \acp{ci}~\cite{christodoulou2024confidence}. 
By approximating unreported uncertainty using a second-order polynomial model of the mean Dice similarity coefficient (DSC), the authors demonstrated that the median width of the reconstructed \acp{ci} was roughly three times larger than the median performance difference between the first- and second-ranked methods.  
This observation aligns with broader concerns in machine learning research, where improper statistical analysis, selective reporting, and the over-interpretation of marginal performance improvements have been identified as key factors limiting reproducibility and the reliability of claims regarding architectural superiority~\cite{pineau2021improving}.

Motivated by these limitations, we conduct a systematic and controlled empirical study to address the following research question:
\textit{How do \acp{gpvm} compare to domain-specific \acp{sma} in 2D \ac{mis} under strictly standardized training and evaluation conditions regarding their segmentation performance?}
Specifically, we evaluate models across three heterogeneous 2D \ac{mis} datasets: (i) RGB binary lesion segmentation, (ii) RGB multi-class polyp segmentation, and (iii) grayscale multi-class cardiac region segmentation. We compare eleven representative architectures, including transformer-based, state-space, and U-Net-derived \acp{sma} as well as recent \ac{gp} \ac{ss} and vision models.
Overall, our contributions are two-fold: 

\textbf{1. Controlled Empirical Study Under a Standardized Benchmarking Protocol.} 
We systematically evaluate \acp{sma} and \acp{gpvm} across three heterogeneous 2D \ac{mis} tasks. 
To mitigate non-architectural sources of variation and confounding, we introduce a strictly controlled benchmarking protocol that enforces consistent training and evaluation procedures across all models, including unified augmentation strategies tailored to each dataset.
In addition, we quantify performance variability, perform statistical comparisons between model families, and provide qualitative insights using \ac{gradcam} attribution maps.

\textbf{2. Practical Insights.} 
Across the evaluated 2D settings, \acp{gpvm} achieved performance largely comparable to that of \acp{sma} under the standardized experimental conditions. In light of recent evidence questioning the statistical robustness of reported performance gains in \ac{mis}~\cite{christodoulou2024confidence}, these findings underscore the importance of rigorous evaluation protocols and informed, resource-aware model selection. 
Overall, we found no compelling evidence of a practically meaningful performance gap between the two model families. \Acp{gpvm} were competitive, but neither family showed a consistent performance advantage or disadvantage.
\section{Benchmarking Methodology}
\label{sec:methodology}

\subsection{Model Selection}
\label{ssec:methodology:model_selection}

Given the rapid growth of (medical) \ac{ss} models and \acp{gpvm}, including all recent methods is impractical. We therefore organize the comparison into two model families, aiming to construct a balanced and representative evaluation, with architectures detailed in Appendix~A: 

\textbf{1. \ac{sma}}: We include five models that are specifically designed for \ac{mis}: \mbox{\acl{hiformer}}~\cite{heidari2023hiformer}, \acl{missformer}~\cite{huang2022missformer}, \acl{swinumamba}~\cite{liuSwinUMambaMICCAI2024}, and CENet~\cite{bozorgpour2025CENet}, alongside the classical \acs{unet}~\cite{ronneberger2015unet} as a widely adopted biomedical baseline.
To ensure a more comparable parameter budget to the selected \acp{gpvm} and to mitigate potential confounding effects due to large differences in model capacity, we adopt stronger encoder variants than those originally proposed for \ac{hiformer} and CENet. Specifically, we replace the ResNet50 backbone in \ac{hiformer} with ResNet101 (denoted \acl{hiformerRN}, or \acs{hiformerRN}), and substitute the PVTv2-B2 encoder in CENet with PVTv2-B3 (\acs{cenet}, i.e., \acl{cenet}). In the remainder of the paper, models marked with an asterisk refer to these enhanced-capacity variants.
Collectively, these architectures span a range of contemporary design paradigms, including convolutional encoder-decoder designs, transformer-based models, hybrid CNN--transformer architectures, and hybrid CNN--state-space models based on Mamba~\cite{gu2024mamba}.

\textbf{2. \ac{gpvm}}: To assess their performance beyond standard vision benchmarks, i.e., for \ac{mis}, we include \acp{gpvm} originally developed for natural-image understanding. This category comprises (i) \ac{ss} architectures and (ii) modern vision backbones (VB) adapted for \ac{ss} using a \acs{uperhead}~\cite{xiao2018uperhead} decoder. 
The former include \acl{segformer}~\cite{xie2021segformer}, \mbox{\acl{segnext}~\cite{guo2022segnext}}, and the \acl{vwformer} decoder~\cite{yan2024vwformer}, which we evaluated with two different backbones, \mbox{MiT-B3} and ConvNeXt-S.
The latter are represented by \acl{internimage}~\cite{wang2023internimage} and \acl{transnext}~\cite{shi2024transnext}.

\begin{table}[ht]
\caption{Overview of the selected methods.}
\label{tab:app:selection:overview}
\centering
\begin{threeparttable}
    \fontsize{8pt}{8pt}\selectfont
    \begin{tabular}{l@{\hskip 0.1in}l@{\hskip 0.2in}c@{\hskip 0.2in}l@{\hskip 0.2in}cc}
        \toprule
        Category & Architecture & Type & Size & Venue & Year \\
        \midrule
        \multirow{5}{*}{\ac{sma}}  
            & \acl{unet}~\cite{ronneberger2015unet}             & CNN                       & 31.0M     & MICCAI        & '15 \\  
            & \acl{hiformerRN}~\cite{heidari2023hiformer}       & Hybrid (CNN-ViT)          & 44.6M     & WACV          & '23 \\   
            & \acl{missformer}~\cite{huang2022missformer}       & Transformer               & 42.5M     & IEEE TMI      & '23 \\   
            & \acl{swinumamba}~\cite{liuSwinUMambaMICCAI2024}   & Hybrid (Mamba-CNN)       & 59.9M     & MICCAI        & '24 \\ 
            & \acl{cenet}~\cite{bozorgpour2025CENet}            & Hybrid (ViT-CNN)          & 53.2M     & MICCAI        & '25 \\
        \midrule
        \multirow{3}{*}{\ac{gpvm} (SS)} 
            & \acl{segformer}~\cite{xie2021segformer}           & Transformer       & 47.2M   & NeurIPS & '21 \\                  
            & \acl{segnext}~\cite{guo2022segnext}               & CNN               & 48.8M   & NeurIPS & '22 \\                  
            & $2\times$ \acl{vwformer}~\cite{yan2024vwformer}  & \multicolumn{2}{c}{--- Backbone-dependent\tnote{1} ---}  & ICLR   & '24  \\   
        \midrule
        \multirow{2}{*}{\ac{gpvm} (VB)}
            & \acl{internimage}~\cite{wang2023internimage}\tnote{2}      & CNN          & 58.3M  & CVPR & '23\\       
            & \acl{transnext}~\cite{shi2024transnext}\tnote{2}           & Transformer  & 57.7M  & CVPR & '24 \\      
        \bottomrule
    \end{tabular}
    \begin{tablenotes}[para]
    \scriptsize{
    \item[1] 51.4M with MiT-B3 (\acs{vwmit}); 57.0M with ConvNeXt-S (\acs{vwconv})
    \item[2] Parameter count includes UPerHead decoder
    }
    \end{tablenotes}
\end{threeparttable}        
\end{table}

Except for \ac{unet} as a widely used baseline, the final selection (cf. Table~\ref{tab:app:selection:overview}) was mainly informed by five criteria:
\begin{enumerate*}[label=(\Roman*)]
\item Architectural diversity across CNN-, \acs{vit}, hybrid-, and emerging paradigms,
\item comparable computational scale (40--60M parameters), either natively or via encoder scaling,
\item scientific visibility in peer-reviewed venues,
\item public code availability to reduce reproduction bias, and
\item availability of ImageNet-pretrained (encoder) weights (either directly provided or obtainable).
\end{enumerate*}

\subsection{Dataset Selection}
\label{ssec:methodology:dataset_selection}

We evaluate all models on three heterogeneous \ac{mis} datasets: \acl{isic}~\cite{codella2019isic,tschandl2018ham10000}, \acl{bkai}~\cite{bkai-igh}, and \acl{camus}~\cite{leclerec2019camus}. As summarized in Table~\ref{tab:dataset_selection}, the datasets differ in imaging modality, class configuration (binary vs. multi-class), and task characteristics, enabling comparison across different imaging settings under controlled conditions.
Preprocessing and data augmentation are adapted to each modality but applied identically across architectures to improve comparability and reduce non-architectural confounding. 
To prevent potential data leakage, we perform dataset-specific filtering procedures that remove duplicate and highly similar images based on identical raw bytes and perceptual hash similarity. Details about data augmentation and image filtering are provided on \href{\githuburl}{GitHub}.

For the main evaluation, we employ five-fold \ac{cv} with approximately equal-sized folds. Random image-level splits are used for \ac{bkai} and \ac{isic}, while patient-aware splitting is applied to \ac{camus}, where patient identifiers are available, to avoid subject-level information leakage.

\begin{table}[ht]
\centering
\caption{Dataset overview ($N$: images after filtering, $C$: classes).}
\label{tab:dataset_selection}
\fontsize{8pt}{8pt}\selectfont
\begin{tabular}{l@{\hskip 0.1in}c@{\hskip 0.1in}c@{\hskip 0.1in}c@{\hskip 0.1in}c@{\hskip 0.15in}c@{\hskip 0.15in}c}
\toprule
Dataset & Modality & Color & $N$ & $C$ & Targets & Characteristic \\
\midrule
\ac{isic}  & Dermoscopy & RGB  & \numprint{3565} & 2 & Lesions         & Irregular boundaries \\
\acs{bkai} & Endoscopy  & RGB  & \numprint{945}  & 3 & Polyps          & Subtype variability; minority class ($C_1$) \\
\ac{camus} & Echocardiography & Gray & \numprint{1996} & 4 & Cardiac regions & Noisy ultrasound data \\
\bottomrule
\end{tabular}
\end{table}

\subsection{Standardized Training and Evaluation Protocol}
\label{ssec:methodology:training_and_eval_unification}

\subsubsection{Training}
All models were trained under a unified protocol while preserving architecture-specific design choices recommended by the original authors or implementations. Standardized settings included ImageNet-pretrained encoders, an input resolution of $256\times256$, the \textit{AdamW} optimizer, the \textit{Reflected Exponential (REX)} learning-rate scheduler~\cite{chen2022rex}, a batch size of 8, and dataset-specific losses: \textit{BCEWithLogitsLoss} (binary) and \textit{CrossEntropyLoss} (multi-class). 

\textbf{Data Augmentation.}
Within each dataset, all architectures used identical, modality-specific augmentation pipelines and normalization based on ImageNet statistics.
For \acs{isic} and \acs{bkai}, extensive affine and photometric augmentations were employed, including, among others, rotations, translations, scalings, intensity jitter, and random flipping to improve robustness against variations in orientation, illumination, and acquisition conditions. For \acs{camus}, more conservative augmentations were adopted. Grayscale images were replicated across three channels, while the extent of affine and photometric perturbations was reduced and flipping was disabled to preserve anatomical plausibility in echocardiography.

\textbf{Experimental Workflow.}
To account for potential differences in learning-rate (LR) sensitivity, we conducted a preliminary study evaluating LRs in $\{10^{-4}, 5\times10^{-5}, 10^{-5}\}$ for each model--dataset pair. Random $60/20/20$ train/validation/test splits were used for \acs{bkai} and \acs{isic}, and the official $80/10/10$ split for \acs{camus}. Each configuration was trained for 100 epochs, and the LR yielding the highest validation Intersection over Union (IoU) was selected for the subsequent five-fold \ac{cv} over the full dataset. Within each fold, models were trained for up to 150 epochs with early stopping (patience 20; minimum validation-IoU improvement $0.0001$), and the checkpoint with the highest validation IoU was evaluated on the corresponding test fold. Hence, each image appeared in one test fold across the \ac{cv}.

\textbf{Implementation Details.}
All models were implemented in PyTorch~2.5.1 (Python~3.11) and trained on two NVIDIA A100 GPUs using mixed-precision training~\cite{micikevicius2018mixed}, with the exception of \acs{segnext}, which exhibited numerical instability under this setting. Deterministic execution was enabled wherever supported to improve reproducibility.
Several architectures required noteworthy adaptations due to implementation constraints. 
In particular, the \acp{sma} \acs{missformer}, \acs{hiformerRN}, and \acs{cenet} were originally designed for fixed input resolutions (e.g., $224 \times 224$) and were not directly compatible with other resolutions such as $256 \times 256$ without modification. 
For \acs{unet}, no pretrained weights were available in the standard implementation; therefore, the first four encoder stages were initialized using ImageNet-pretrained VGG13 convolutional blocks. 
For \acs{vwformer}, missing configuration details in the original code repository were resolved by setting \texttt{nheads}=1 and enabling shortcut connections.

\subsubsection{Evaluation}

Segmentation performance was measured using IoU, DSC, recall, and precision. 
All metrics were calculated exclusively over foreground classes, with the background class excluded from evaluation.
Although IoU was employed as the early stopping criterion, DSC served as the primary reporting metric due to its widespread adoption in \acs{mis}. Since IoU and DSC are monotonically related through $DSC = 2 IoU/(1+IoU)$ under identical contingency counts, optimizing IoU during training is consistent with DSC-based evaluation.

\textbf{Metrics Calculation.}
For each dataset and model, performance was computed using \textit{globally pooled micro-averaging}. Specifically, true positives (TP), false positives (FP), and false negatives (FN) were aggregated across all test images and foreground classes. Metrics were then derived from these global counts. DSC is defined as \(2TP/(2TP+FP+FN)\), IoU as \(TP/(TP+FP+FN)\), recall as \(TP/(TP+FN)\), and precision as \(TP/(TP+FP)\). 
For class-wise evaluation, confusion counts were aggregated separately per foreground class before metric computation. 
For fold-wise analysis, the same aggregation procedure was applied within each fold, yielding fold-level metrics based on pooled counts. 
Image-level class-wise DSC was computed for all images in which the class was present in the reference or the prediction, and used for visualization of fold-level variability (cf. Fig.~C1~--~C3). 
Overall, this evaluation strategy avoids dominance of the background while providing a consistent global micro-averaged estimate of segmentation performance across foreground regions.

\textbf{Uncertainty Estimates.}
Uncertainty of the pooled estimates, i.e., global and class-wise metric scores, was quantified using a non-parametric, image-level bootstrap procedure (stratified by \ac{cv} folds) to account for image-to-image variability while avoiding treating spatially correlated pixels as independent observations. 
Furthermore, the bootstrap method imposes no distributional assumptions on either the metric or the sampling distribution of the mean~\cite{el2025confidence}.
For each dataset–architecture pair, images were resampled with replacement within each of the five folds, preserving the original fold sizes. For each of the 2000 bootstrap replicates, foreground confusion counts were re-aggregated over the resampled images, and metrics were recomputed from the resulting pooled counts. The reported 95\% \acp{ci} correspond to the 2.5th and 97.5th percentiles of the bootstrap distribution, while the bootstrap standard deviation (bSD) reflects variability of the pooled estimator. The same procedure was applied for the class-specific pooled DSC estimates, but calculated separately per class.

\textbf{Statistics.}
Statistical analysis was conducted in JASP~\cite{JASP2026}, version 0.97.0, and alpha was set at $.05$.
To analyze the data, linear mixed-effects models (LMMs) were used, with image-level micro-averaged DSC as the dependent variable, \textit{ModelFamily} (either \ac{sma} or \ac{gpvm}) as fixed effect, and random intercepts for \textit{Dataset}, \textit{Fold}, and \textit{Model} to account for the hierarchical structure of the data (i.e., repeated segmentation of images across datasets, folds, and models). 
Specifically, a separate micro-averaged DSC score was computed for each image, dataset, and architecture by first pooling TP, FP, and FN over all foreground classes within that image and then computing DSC from these image-level pooled counts. The resulting DSC scores per image-architecture-pair served as input for the LMM.

Model adequacy and complexity were assessed using the Bayesian Information Criterion (BIC), based on the discussion in Aho et al.~\cite{aho2014}, and Likelihood Ratio Tests (LRTs), to compare how well the same dataset (in relation) supports the more complex model versus the simpler model~\cite{meteyard2020}. Table~\ref{tab:lmm-comparison-selection} depicts an overview of models that were fit to the data, including a null model as baseline, and compared against each other using LRTs. 
Model comparison and selection are reported following the recommendations of Meteyard and Davies~\cite{meteyard2020}.

\begin{table}[ht]
\centering
\caption{Linear mixed-effects models fitted to the data for this study.}
\label{tab:lmm-comparison-selection}

\fontsize{8pt}{8pt}\selectfont

\begin{tabular}{l@{\hskip 0.1in}c@{\hskip 0.1in}ccc@{\hskip 0.15in}cccc@{\hskip 0.15in}cc}
\toprule

\multirow{2}{*}{Model} & \multirow{2}{*}{Fixed effects}  &
\multicolumn{3}{c@{\hskip 0.15in}}{Random effects} &
\multicolumn{2}{c@{\hskip 0.15in}}{Model fit} &
\multicolumn{2}{c}{LRT} \\

\cmidrule(lr){3-5} \cmidrule(lr){6-7} \cmidrule(lr){8-9}

 &  & Dataset & Fold & Model &
 BIC & df &
df & $\chi^2$ \\
\midrule

Null model  & - & I  & I & I &
 -68130 &  5 &
 &  \\

FE model & ModelFamily  & I  & I & I &
 -68130 &  6 &
1 & 7.42* \\

RS model & ModelFamily & I + RS & I + RS & I + RS &
 -68210  & 10 &
1 & 2.23 \\

\bottomrule
\end{tabular}

\vspace{0.2cm}
\footnotesize{\textit{Note.} LRT = Likelihood Ratio Test, BIC = Bayesian Information Criterion, df = degrees of freedom, \\ 
I = intercepts, FE = fixed effects, RS = random slopes. * $p < .05$}
\end{table}

As suggested by Barr et al.~\cite{barr2013}, we initially opted for a maximal random-effect structure with the random-slopes model (RS model), thereby allowing the effect of \textit{ModelFamily} to vary across datasets. However, this model did not converge properly and resulted in a singular fit, with near-zero variance estimates ($\sigma^2 \approx 0$) for the random slope component. This indicates that the data did not support the additional complexity of allowing dataset-specific variation in the \textit{ModelFamily} effect. After successively removing random slopes, this resulted in a parsimonious random-effects structure (FE model), which is defined as follows:

\begin{equation}
    DSC = \beta_0 + \beta_1 \,\text{\textit{ModelFamily}} + u_{\text{Dataset}} + u_{\text{Fold}} + u_{\text{Model}} + \varepsilon,
\end{equation}

where $\beta_0$ denotes the overall intercept corresponding to the expected DSC of the reference group when all random effects are zero, and $\beta_1$ represents the fixed effect of \textit{ModelFamily}, quantifying the difference in expected DSC between \ac{sma} and \ac{gpvm}. The terms $u_{Dataset}$, $u_{Fold}$, and $u_{Model}$ are random intercepts capturing \textit{Dataset}-specific, \textit{Fold}-specific, and \textit{Model}-specific deviations from the global mean, respectively, thereby accounting for hierarchical sources of variability across the experimental design.
In particular, $u_{Model}$ explicitly accounts for variability between individual models within each model family.
Finally, $\epsilon$ denotes the residual error term, representing unexplained variation at the image level.

\textbf{Computational Efficiency and Explainability.}
Computational efficiency was quantified using parameter count and \acsp{gmac}, estimated via the \texttt{calflops} library~\cite{calflops2023} for an input shape of $(1, 3, 256, 256)$. GFLOPs were not reported separately, as they can be derived from GMACs through a fixed conversion factor.
Model interpretability was evaluated using \ac{gradcam} visualizations on selected test samples.
The heatmaps were generated with a modified \texttt{M3d-CAM} implementation~\cite{m3dCAM2020} using \texttt{layer='auto'}, which automatically selects the last suitable layer for extracting class-discriminative activation maps.

\section{Evaluation}
\label{sec:eval}

\subsection{Segmentation Performance}
\label{ssec:eval:segmentation_performance}

Table~\ref{tab:eval:per_dataset_metrics} reports foreground-only, globally pooled micro-averaged estimates for each dataset. 
Global DSC serves as the primary metric, complemented by IoU, recall, precision, and class-wise DSC. All metrics are reported with the bootstrap standard deviation (bSD) of the pooled estimator. Importantly, this uncertainty reflects variability under image-level resampling; it should not be interpreted as fold-level variability or as image-level heterogeneity.

\begin{figure*}[ht]
\begin{adjustbox}{max width=\textwidth, keepaspectratio}
\begin{tabular}{cc}
\bmvaHangBox{\includegraphics[trim=0 0 5 23, clip, width=0.5\textwidth]{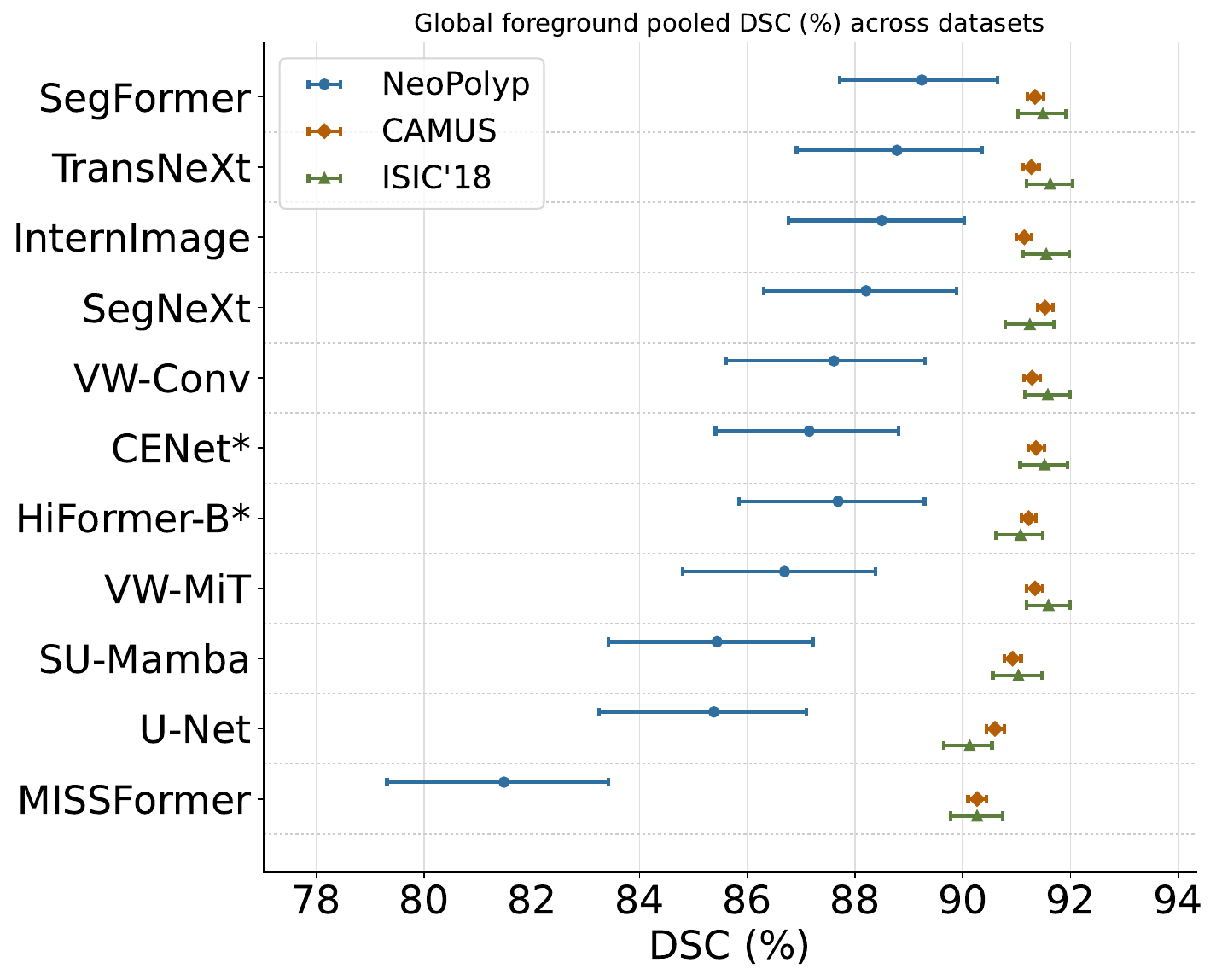}} &
\bmvaHangBox{\includegraphics[trim=0 0 5 23, clip, width=0.5\textwidth]{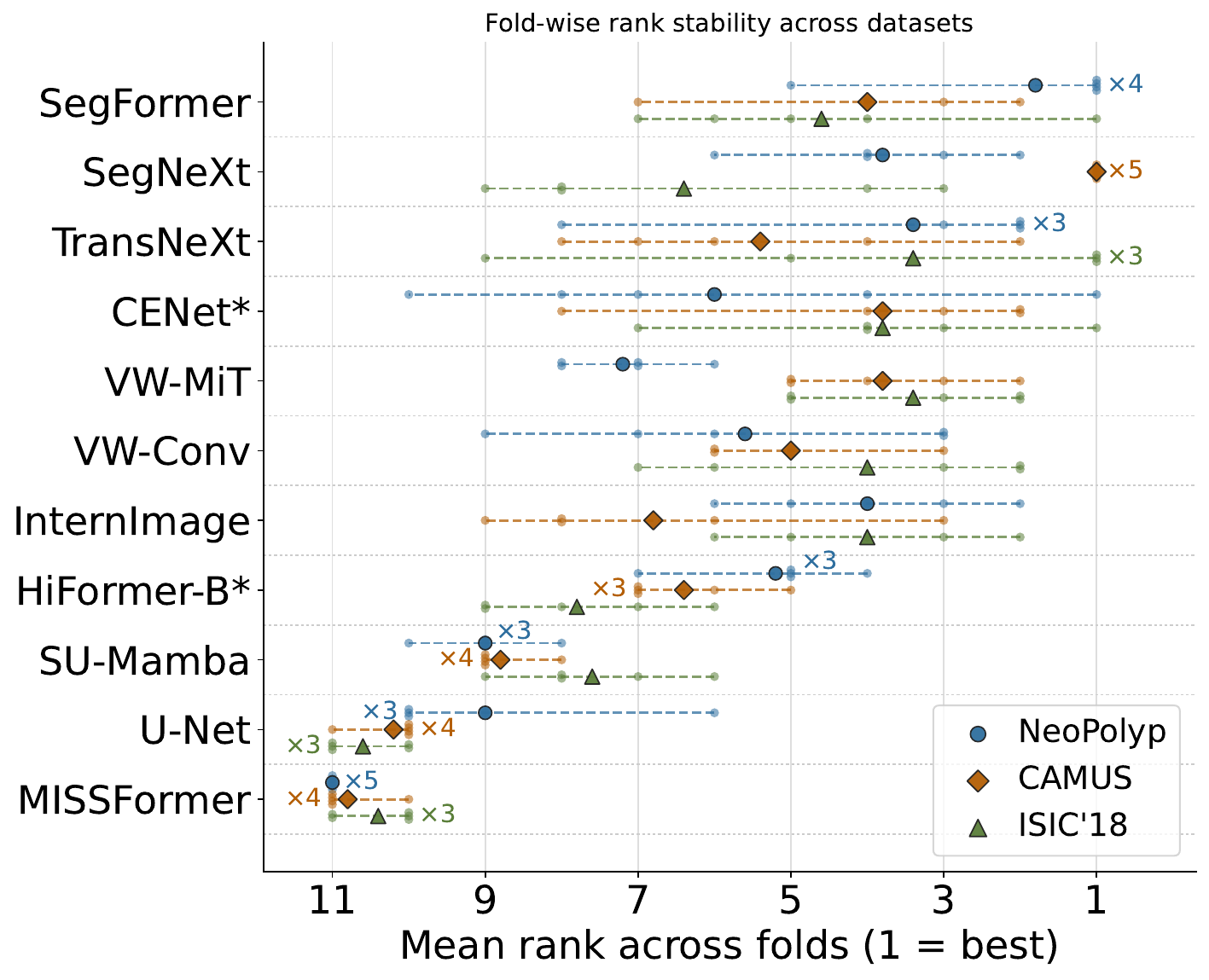}}\\
(a) Global micro-averaged DSC & (b) Rank stability across folds
\end{tabular}
\end{adjustbox}
\vspace{\abovecaptionskip}
\caption{Global segmentation performance and cross-fold rank stability.}
\label{fig:global_perspective}
\end{figure*}

\begin{figure}[ht]
\begin{adjustbox}{max width=\textwidth, keepaspectratio}
\begin{tabular}{cc}
\bmvaHangBox{\includegraphics[trim=0 0 5 23, clip, width=0.5\textwidth]{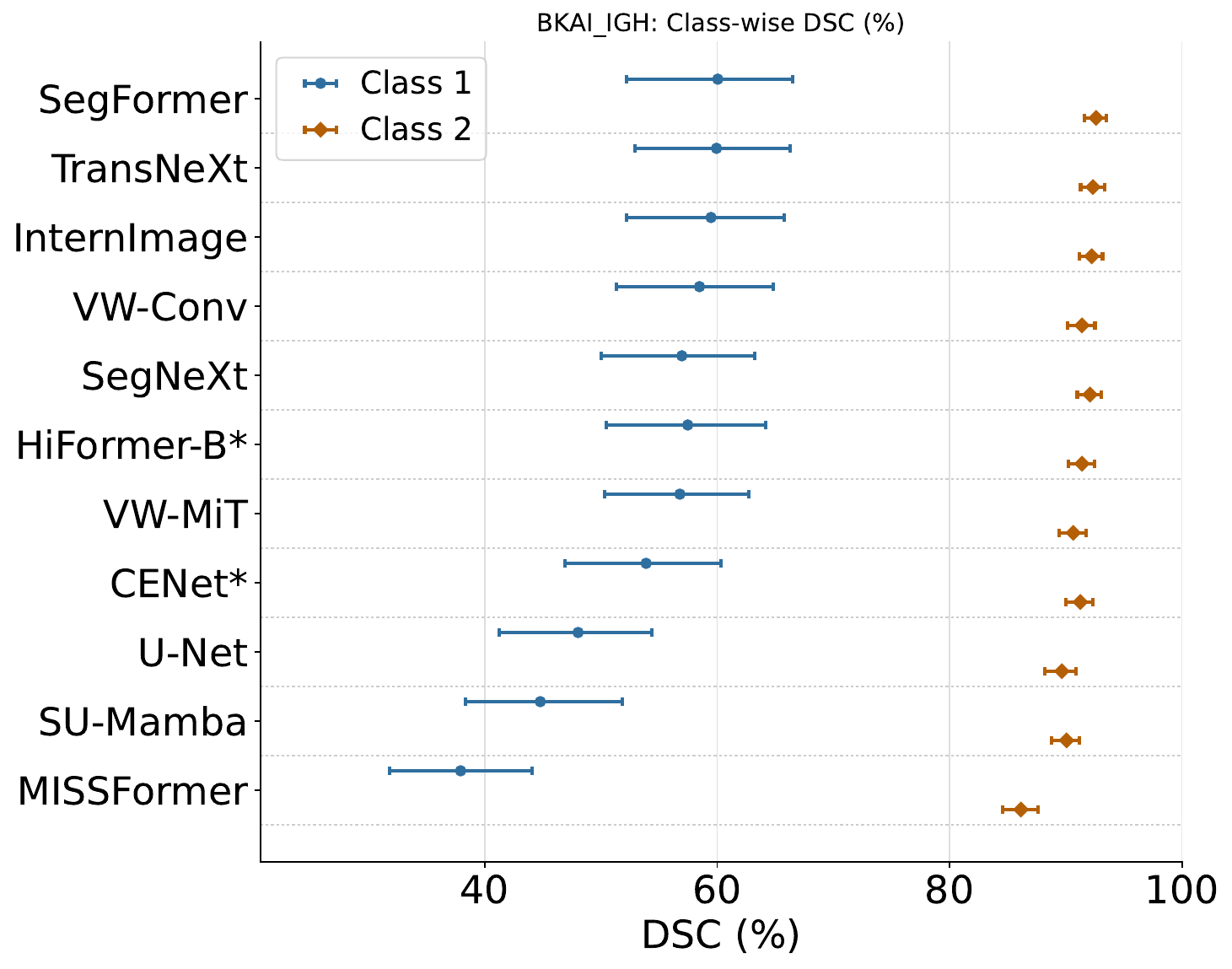}} &
\bmvaHangBox{\includegraphics[trim=0 0 5 23, clip, width=0.5\textwidth]{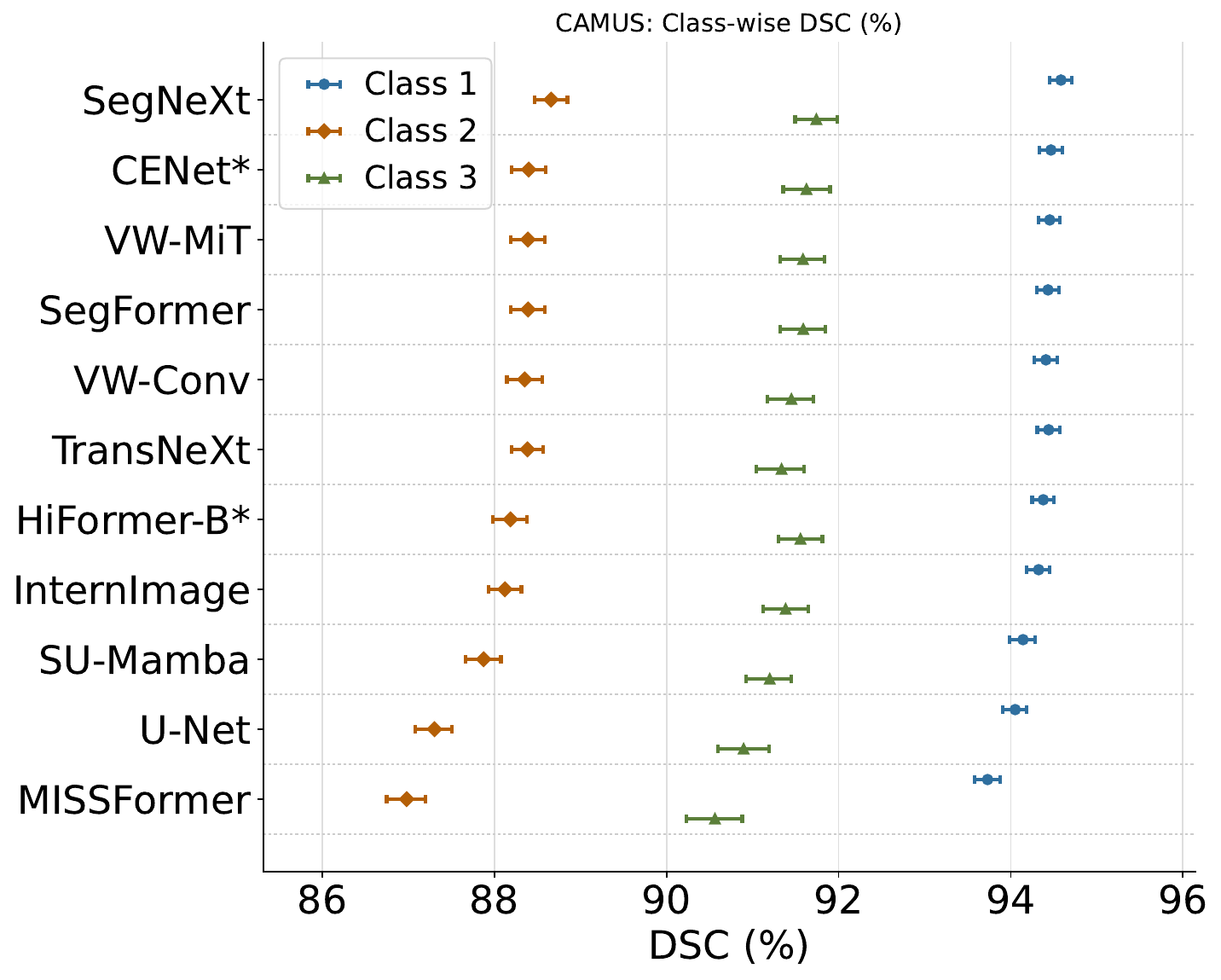}}\\
(a) NeoPolyp & (b) CAMUS
\end{tabular}
\end{adjustbox}
\vspace{\abovecaptionskip}
\caption{Class-wise global DSC for the two multi-class datasets.}
\label{fig:classwise_perspective}
\end{figure}

Figure~\ref{fig:global_perspective}(a) visualizes the global micro-averaged DSC for each architecture and dataset, with horizontal bars indicating bootstrap-based 95\% \acp{ci}. 
Complementarily, Figure~\ref{fig:classwise_perspective} presents class-wise DSC values for each method on the two multi-class datasets, NeoPolyp and CAMUS, with corresponding bootstrap-based 95\% \acp{ci}. 
Finally, the rank stability plot in Figure~\ref{fig:global_perspective}(b) summarizes the relative performance of each method across the five \ac{cv} folds. Within each fold, models are ranked by global micro-DSC. 
A dashed horizontal line spans from the minimum to the maximum observed rank, with individual fold ranks plotted along the line and the mean rank marked by a dataset-specific marker. Ranks occurring in three or more folds are annotated with their frequency. Thus, wider lines indicate greater variability in relative rankings across folds and hence lower stability of model ordering across splits.

\begin{table}[ht!]
\centering
\caption{Per-dataset global micro-averaged metrics (pooled estimate ± bSD, in \%).}
\label{tab:eval:per_dataset_metrics}
\begin{threeparttable}
    \fontsize{8pt}{8pt}\selectfont
    \begin{tabular}{l@{\hskip 0.15in} 
        l@{\hskip 0.1in}
        c@{\hskip 0.05in}c@{\hskip 0.05in}c@{\hskip 0.05in}c@{\hskip 0.15in}
        c@{\hskip 0.05in}c@{\hskip 0.05in}c@{\hskip 0.1in}}
        \toprule
        \multirow{2}{*}{} & 
        \multirow{2}{*}{Model} &
        \multirow{2}{*}{DSC} &
        \multirow{2}{*}{IoU} &
        \multirow{2}{*}{Recall} &
        \multirow{2}{*}{Precision} &
        \multicolumn{3}{c}{Class-wise DSC \tnote{1}}\\
        \cmidrule(lr{2pt}){7-9}
        
         & & & & & &  $C_1$ & $C_2$   &$C_3$\\ \midrule\midrule

        \multirow{11}{*}{\rotatebox{90}{\textbf{\ac{bkai}}}}
         & \acs{unet}          & 85.4$\pm$1.0 & 74.5$\pm$1.5 & 83.7$\pm$1.3 & 87.2$\pm$0.9         & 48.0$\pm$3.3 & 89.7$\pm$0.7 & --\\    
         & \acs{hiformerRN}    & 87.7$\pm$0.9 & 78.1$\pm$1.4 & 86.5$\pm$1.1 & 88.9$\pm$0.9         & 57.5$\pm$3.5 & 91.4$\pm$0.6 & --\\    
         & \acs{missformer}    & 81.5$\pm$1.1 & 68.7$\pm$1.5 & 79.1$\pm$1.4 & 84.0$\pm$1.1         & 37.9$\pm$3.2 & 86.1$\pm$0.8 & --\\    
         & \acs{swinumamba}    & 85.4$\pm$1.0 & 74.6$\pm$1.5 & 83.3$\pm$1.2 & 87.7$\pm$0.9         & 44.8$\pm$3.5 & 90.1$\pm$0.6 & --\\    
         & \acs{cenet}         & 87.1$\pm$0.9 & 77.2$\pm$1.4 & 84.9$\pm$1.2 & 89.5$\pm$0.8         & 53.9$\pm$3.4 & 91.2$\pm$0.6 & --\\    
         \cmidrule(lr{2pt}){2-9}
         & \acs{segformer}     & 89.2$\pm$0.7 & 80.6$\pm$1.2 & 87.5$\pm$1.1 & 91.1$\pm$0.6         & 60.1$\pm$3.6 & 92.6$\pm$0.5 & --\\    
         & \acs{segnext}       & 88.2$\pm$0.9 & 78.9$\pm$1.4 & 87.2$\pm$1.2 & 89.2$\pm$0.8         & 57.0$\pm$3.5 & 92.1$\pm$0.5 & --\\    
         & \acs{vwconv}        & 87.6$\pm$0.9 & 77.9$\pm$1.5 & 85.8$\pm$1.2 & 89.5$\pm$0.8         & 58.5$\pm$3.4 & 91.4$\pm$0.6 & --\\    
         & \acs{vwmit}         & 86.7$\pm$0.9 & 76.5$\pm$1.4 & 84.9$\pm$1.2 & 88.6$\pm$0.8         & 56.8$\pm$3.2 & 90.6$\pm$0.6 & --\\    
         & \acs{internimage}   & 88.5$\pm$0.8 & 79.4$\pm$1.3 & 87.2$\pm$1.1 & 89.8$\pm$0.7         & 59.5$\pm$3.4 & 92.2$\pm$0.5 & --\\    
         & \acs{transnext}     & 88.8$\pm$0.9 & 79.8$\pm$1.4 & 87.4$\pm$1.2 & 90.2$\pm$0.8         & 59.9$\pm$3.5 & 92.3$\pm$0.5 & --\\

        \midrule\midrule
        
        \multirow{11}{*}{\rotatebox{90}{\textbf{\ac{camus}}}}
        & \acs{unet}            & 90.6$\pm$0.1 & 82.8$\pm$0.1 & 90.6$\pm$0.1 & 90.6$\pm$0.1         & 94.1$\pm$0.1 & 87.3$\pm$0.1 & 90.9$\pm$0.2 \\
        & \acs{hiformerRN}      & 91.2$\pm$0.1 & 83.9$\pm$0.1 & 91.3$\pm$0.1 & 91.2$\pm$0.1         & 94.4$\pm$0.1 & 88.2$\pm$0.1 & 91.6$\pm$0.1 \\
        & \acs{missformer}      & 90.3$\pm$0.1 & 82.3$\pm$0.1 & 90.5$\pm$0.1 & 90.1$\pm$0.1         & 93.7$\pm$0.1 & 87.0$\pm$0.1 & 90.6$\pm$0.2 \\
        & \acs{swinumamba}      & 90.9$\pm$0.1 & 83.4$\pm$0.1 & 91.0$\pm$0.1 & 90.9$\pm$0.1         & 94.1$\pm$0.1 & 87.9$\pm$0.1 & 91.2$\pm$0.1 \\
        & \acs{cenet}           & 91.4$\pm$0.1 & 84.1$\pm$0.1 & 91.3$\pm$0.1 & 91.4$\pm$0.1         & 94.5$\pm$0.1 & 88.4$\pm$0.1 & 91.6$\pm$0.1 \\
        
        \cmidrule(lr{2pt}){2-9}

        & \acs{segformer}       & 91.3$\pm$0.1 & 84.1$\pm$0.1 & 91.2$\pm$0.1 & 91.5$\pm$0.1         & 94.4$\pm$0.1 & 88.4$\pm$0.1 & 91.6$\pm$0.1 \\
        & \acs{segnext}         & 91.5$\pm$0.1 & 84.4$\pm$0.1 & 91.5$\pm$0.1 & 91.6$\pm$0.1         & 94.6$\pm$0.1 & 88.7$\pm$0.1 & 91.7$\pm$0.1 \\
        & \acs{vwconv}          & 91.3$\pm$0.1 & 84.0$\pm$0.1 & 91.3$\pm$0.1 & 91.3$\pm$0.1         & 94.4$\pm$0.1 & 88.4$\pm$0.1 & 91.5$\pm$0.1 \\
        & \acs{vwmit}           & 91.3$\pm$0.1 & 84.1$\pm$0.1 & 91.4$\pm$0.1 & 91.3$\pm$0.1         & 94.5$\pm$0.1 & 88.4$\pm$0.1 & 91.6$\pm$0.1 \\
        & \acs{internimage}     & 91.1$\pm$0.1 & 83.7$\pm$0.1 & 91.0$\pm$0.1 & 91.3$\pm$0.1         & 94.3$\pm$0.1 & 88.1$\pm$0.1 & 91.4$\pm$0.1 \\
        & \acs{transnext}       & 91.3$\pm$0.1 & 83.9$\pm$0.1 & 91.4$\pm$0.1 & 91.1$\pm$0.1         & 94.4$\pm$0.1 & 88.4$\pm$0.1 & 91.3$\pm$0.1 \\
        
        \midrule\midrule
        
        \multirow{11}{*}{\rotatebox{90}{\textbf{\ac{isic}}}}
         & \acs{unet}           & 90.1$\pm$0.2 & 82.0$\pm$0.4 & 89.3$\pm$0.4 & 91.0$\pm$0.3     & --  & --  & --\\
         & \acs{hiformerRN}     & 91.1$\pm$0.2 & 83.6$\pm$0.4 & 90.4$\pm$0.4 & 91.8$\pm$0.3     & --  & --  & --\\
         & \acs{missformer}     & 90.3$\pm$0.2 & 82.3$\pm$0.4 & 89.2$\pm$0.4 & 91.4$\pm$0.3     & --  & --  & --\\
         & \acs{swinumamba}     & 91.0$\pm$0.2 & 83.5$\pm$0.4 & 90.0$\pm$0.4 & 92.1$\pm$0.3     & --  & --  & --\\
         & \acs{cenet}          & 91.5$\pm$0.2 & 84.4$\pm$0.4 & 90.1$\pm$0.4 & 93.0$\pm$0.3     & --  & --  & --
         \\\cmidrule(lr{2pt}){2-6}
         & \acs{segformer}      & 91.5$\pm$0.2 & 84.3$\pm$0.4 & 90.7$\pm$0.4 & 92.3$\pm$0.3     & --  & --  & --\\
         & \acs{segnext}        & 91.2$\pm$0.2 & 83.9$\pm$0.4 & 90.9$\pm$0.3 & 91.6$\pm$0.3     & --  & --  & --\\
         & \acs{vwconv}         & 91.6$\pm$0.2 & 84.5$\pm$0.4 & 90.5$\pm$0.4 & 92.7$\pm$0.3     & --  & --  & --\\
         & \acs{vwmit}          & 91.6$\pm$0.2 & 84.5$\pm$0.4 & 90.2$\pm$0.4 & 93.0$\pm$0.3     & --  & --  & --\\
         & \acs{internimage}    & 91.6$\pm$0.2 & 84.4$\pm$0.4 & 90.8$\pm$0.4 & 92.4$\pm$0.3     & --  & --  & --\\
         & \acs{transnext}      & 91.6$\pm$0.2 & 84.5$\pm$0.4 & 90.9$\pm$0.3 & 92.4$\pm$0.3     & --  & --  & --\\
        
        \bottomrule
    
    \end{tabular}
    \begin{tablenotes}[para]
    \scriptsize{
    \item[1] \ac{bkai}: $C_1$ - non-neoplastic; $C_2$ - neoplastic | \ac{camus}: $C_1$ - LV Cavity; $C_2$ - LV Wall; $C_3$ - LA
    }
    \end{tablenotes}
\end{threeparttable}        
\end{table}

\textbf{Descriptive DSC estimates are closely clustered across architectures.}
Across the three datasets, pooled DSC estimates indicate broadly similar segmentation performance of the evaluated models. For descriptive cross-dataset comparisons, we additionally computed the unweighted arithmetic mean of the three dataset-level pooled DSC estimates for each model. 
Several \acp{gpvm} achieved the highest values, with \acs{segformer} reaching 90.7\%, followed by \acs{transnext} at 90.6\%, and \acs{internimage} and \acs{segnext} at 90.4\% and 90.3\%, respectively. Among the \acp{sma}, \acs{cenet} and \mbox{\acs{hiformerRN}} reached the highest means, both at 90.0\%. Thus, the descriptive summary shows a small numerical tendency toward higher values for \acp{gpvm}, although differences between the strongest architectures remain small.

\textbf{The spread of pooled DSC estimates differs across datasets.}
The global micro-DSC plot in Figure~\ref{fig:global_perspective}(a) shows the greatest separation between models on \ac{bkai},
where both the range of point estimates and the associated variability are largest.  
In contrast, \ac{camus} and \ac{isic} exhibit tighter clustering of global micro-DSC values, mostly located in a narrow performance range around 91\%. This indicates that descriptive differences between architectures are more pronounced in the \ac{bkai} setting than in the other two datasets.

\textbf{Class-wise plots identify dataset-specific challenges.}
The class-wise, globally pooled DSC results (cf. Table~\ref{tab:eval:per_dataset_metrics}) and Figure~\ref{fig:classwise_perspective} indicate that the increased variability on \ac{bkai} was primarily driven by class $C_1$ (non-neoplastic polyps). This class consistently exhibited lower DSC values and greater variability across models than class $C_2$. In contrast, $C_2$ was segmented more reliably, with DSC scores generally reaching or exceeding 90\% across most architectures. This behavior is consistent with the marked class imbalance: $C_1$ comprises approximately 0.5\% of all pixels, compared with roughly 3.5\% for $C_2$.
For \ac{camus}, inter-class differences are less pronounced. The left ventricular (LV) wall ($C_2$) consistently yielded lower DSC values than the LV cavity ($C_1$), while the left atrium (LA, $C_3$) lies between them in terms of class-wise DSC. Regarding class distribution, $C_1$ accounts for approximately 7.6\% of all pixels, $C_2$ for around 8.4\%, and $C_3$ for approximately 4.2\%.
Figure~B1 illustrates the class imbalance for all datasets, particularly relative to the dominant background class.

\textbf{Fold-wise rankings vary across datasets.}
Several \mbox{\acp{gpvm}} consistently ranked among the top five models on individual datasets, including \acs{segformer} on \ac{bkai}, \acs{segnext} on \mbox{\ac{camus}}, and \acs{vwmit} on \ac{isic}. In contrast, \acs{missformer}, \acs{unet}, and \acs{swinumamba} often occupied the lowest ranks. Averaged within model families, \acp{gpvm} achieved a mean rank of $\approx 4.3$ across datasets (\ac{bkai}: $4.30$, \ac{camus}: $4.33$, \ac{isic}: $4.30$), compared with $\approx 8.0$ for \acp{sma} ($8.04$, $8.00$, and $8.04$, respectively).
Except for \acs{segnext} on \ac{camus} and \acs{missformer} on \ac{bkai}, which ranked first and last, respectively, in every fold, rankings exhibited relatively wide, dataset-dependent intervals, indicating limited stability in model ordering. Thus, small differences in globally pooled DSC point estimates should be interpreted cautiously, as they may be sensitive to data splits and datasets. 
Moreover, because ranks capture only the ordinal ordering of DSC values rather than their magnitude, differences in mean rank do not necessarily indicate substantial differences in performance.

\textbf{Descriptive summary.}
Overall, the table and plots show broadly comparable pooled DSC scores across architectures. The highest numerical point estimates were observed for several \acp{gpvm}, but differences to the strongest \acp{sma} were small. The most prominent descriptive patterns are dataset- and class-dependent: \ac{bkai}, especially class $C_1$, shows the greatest spread, whereas \ac{camus} and \ac{isic} exhibit more compact performance profiles. 

\subsection{Statistical Evaluation Results}
Accounting for dataset, fold, and model, the \ac{sma} model family demonstrated slightly higher DSC performance than the \ac{gpvm} group ($\beta = 0.006, SE = 0.002, t = 3.32$), indicating \textit{ModelFamily} to be a significant predictor of DSC performance ($p = .008$). Since \ac{gpvm} was used as the reference category, the positive regression coefficient indicates a slightly higher adjusted DSC for \ac{sma} compared to \ac{gpvm} after accounting for variability due to dataset, fold, and specific segmentation model. However, the estimated effect size was very small ($d = 0.04$), suggesting only a minor practical difference between model families.

Importantly, the significance of the fixed effect might be affected by the simplified random-effects structure due to singularity and negligible slope variance, potentially resulting in an inflation of Type I error (i.e., increased probability of falsely detecting an effect)~\cite{barr2013}. Thus, this result should be interpreted cautiously.
An overview of all relevant parameter estimates of the LMM can be found in Table~\ref{tab:lmm-estimates}.

\begin{table}[ht]
\centering
\caption{Linear mixed-effects model with parameter estimates for fixed and random effects.}
\label{tab:lmm-estimates}
\fontsize{8pt}{8pt}\selectfont
\begin{tabular}{l@{\hskip 0.1in}c@{\hskip 0.1in}c@{\hskip 0.1in}c@{\hskip 0.1in}c}
\toprule
\textbf{Fixed effects} & $\beta$ & \textit{SE} & \textit{t} & \textit{p} \\
\midrule
Intercept  & 0.862 & 0.035  & 24.85 & $< .001$ \\
ModelFamily & 0.006  & 0.002  & 3.32  & .008  \\
\midrule
\midrule
\textbf{Random effects} & Variance & Standard Deviation &  &  \\
\midrule
Dataset (Intercept) & 0.004 & 0.06 &  &   \\
Fold (Intercept) & 0.000 & 0.004 &  &   \\
Model (Intercept) & 0.000 & 0.006 &  &   \\
\bottomrule
\end{tabular}

\footnotesize{\textit{Note.} p-values for the fixed effects were calculated using Likelihood Ratio Tests.}
\end{table}

\subsection{Computational Efficiency}

\begin{figure*}[ht!]
    \centering
    \includegraphics[width=\textwidth,trim={0cm 10 0 0},clip]{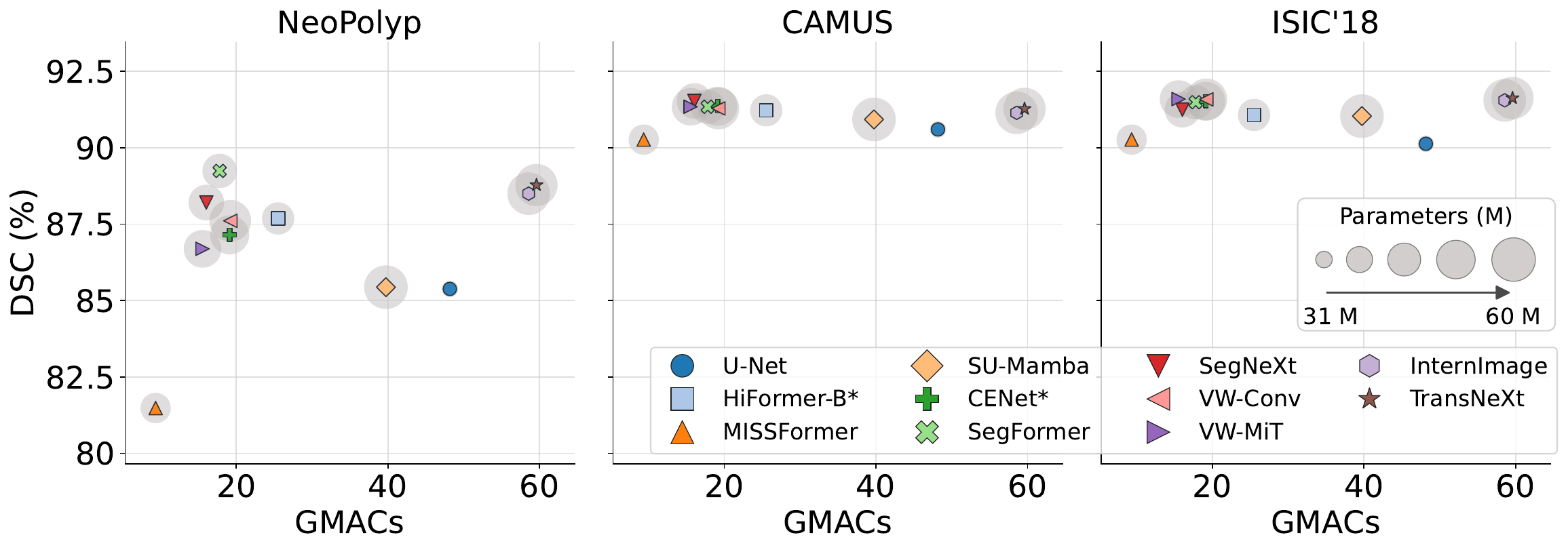}
    \caption{Computational efficiency in terms of \acp{gmac} vs. global micro-averaged DSC. 
   }
    \label{fig:computational_complexity}
\end{figure*}

Figure~\ref{fig:computational_complexity} provides a descriptive comparison of DSC point estimates and computational cost in terms of \acp{gmac}, with gray circle sizes indicating parameter count.
Across datasets, architectures spanned a wide computational range (9.4–59.6 \acp{gmac}), while inter-dataset differences due to varying numbers of output-head channels were negligible (<0.14\%).
Global micro-DSC estimates were comparatively close for \ac{camus} (90.3--91.5\%) and \ac{isic} (90.1--91.6\%), but more dispersed for \ac{bkai} (81.5--89.2\%). 
The highest DSC point estimate was achieved by different architectures across datasets: \acs{segformer} on \ac{bkai} (17.8 \acp{gmac}), \acs{segnext} on \ac{camus} (16.0 \acp{gmac}), and \acs{transnext} on \ac{isic} (59.6 \acp{gmac}). 
These point estimates should, however, be interpreted in conjunction with their associated uncertainty rather than as definitive evidence of superiority.
Overall, the results do not indicate a clear monotonic association between computational cost and DSC. Notably, \ac{missformer} consistently exhibited the lowest computational cost and also ranked among the lower-performing models in terms of DSC, particularly on \ac{bkai}.
More broadly, several architectures achieved comparable DSC despite appreciable differences in \acp{gmac}, supporting the separate consideration of predictive performance and efficiency.

\subsection{Explainability Insights}
\label{ssec:eval:XAI}
\begin{figure*}[ht!]

\centering
\setlength{\tabcolsep}{2pt}
\begin{adjustbox}{max height=\textwidth, max width=0.64\textheight, keepaspectratio}
\begin{tabular}{l c c | c c c | c c c c }

\raisebox{1\height}{\rotatebox[origin=c]{90}{\scriptsize{GT}}} & 

\multicolumn{2}{c|}{\includegraphics[trim=50 50 50 50, clip, width=0.1\textwidth]{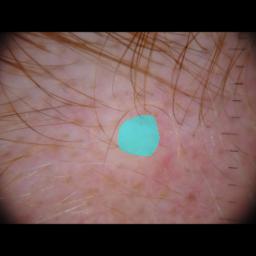}} &
\multicolumn{3}{c|}{\includegraphics[trim=40 36 25 29, clip, width=0.1\textwidth]{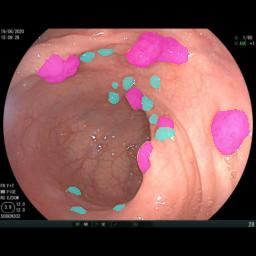}} &
\multicolumn{4}{c}{\includegraphics[trim=40 30 30 40, clip, width=0.1\textwidth]{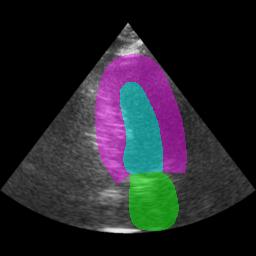}} \\

& Pred. & $C_1$ & Pred.  & $C_1$ & $C_2$ & Pred. & $C_1$& $C_2$& $C_3$ \\

\raisebox{1.2\height}{\rotatebox[origin=c]{90}{\scriptsize{\ac{unet}}}} & 
\includegraphics[trim=50 50 50 50, clip, width=0.1\textwidth]{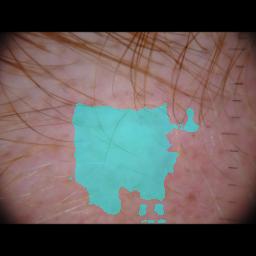} &     
\includegraphics[trim=50 50 50 50, clip, width=0.1\textwidth]{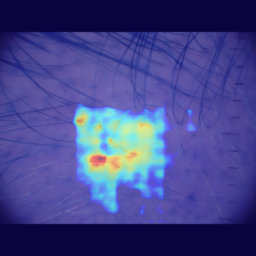} &     
\includegraphics[trim=40 36 25 29, clip, width=0.1\textwidth]{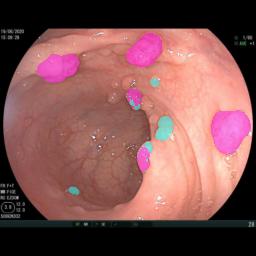} &       
\includegraphics[trim=40 36 25 29, clip, width=0.1\textwidth]{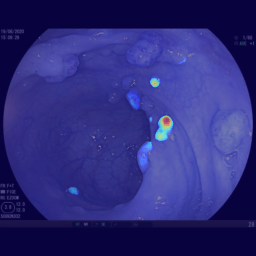} &     
\includegraphics[trim=40 36 25 29, clip, width=0.1\textwidth]{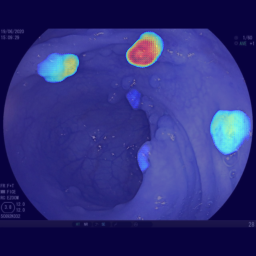} &     
\includegraphics[trim=40 30 30 40, clip, width=0.1\textwidth]{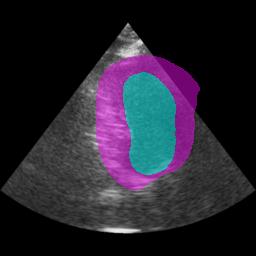} &             
\includegraphics[trim=40 30 30 40, clip, width=0.1\textwidth]{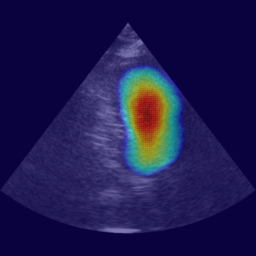} &      
\includegraphics[trim=40 30 30 40, clip, width=0.1\textwidth]{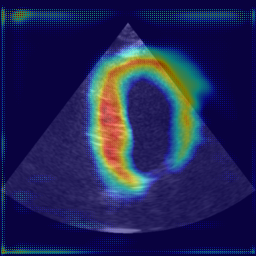}  &     
\includegraphics[trim=40 30 30 40, clip, width=0.1\textwidth]{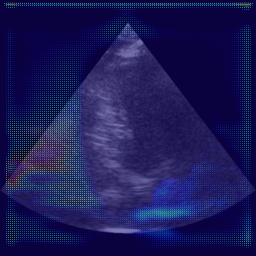}   \\   

\raisebox{0.9\height}{\rotatebox[origin=c]{90}{\scriptsize{\acs{hiformerRN}}}} & 
\includegraphics[trim=50 50 50 50, clip, width=0.1\textwidth]{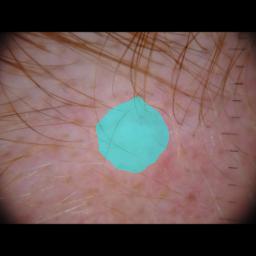} &       
\includegraphics[trim=50 50 50 50, clip, width=0.1\textwidth]{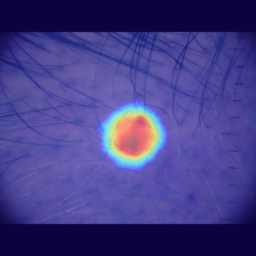} &        
\includegraphics[trim=40 36 25 29, clip, width=0.1\textwidth]{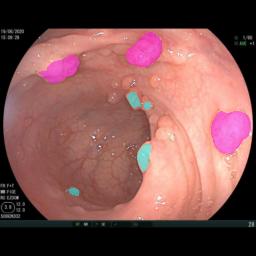} &       
\includegraphics[trim=40 36 25 29, clip, width=0.1\textwidth]{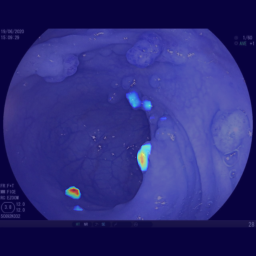} &     
\includegraphics[trim=40 36 25 29, clip, width=0.1\textwidth]{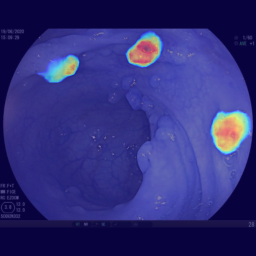} &     
\includegraphics[trim=40 30 30 40, clip, width=0.1\textwidth]{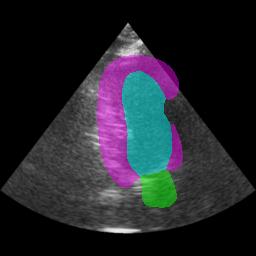} &             
\includegraphics[trim=40 30 30 40, clip, width=0.1\textwidth]{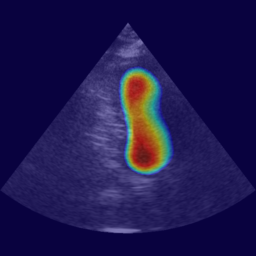} &      
\includegraphics[trim=40 30 30 40, clip, width=0.1\textwidth]{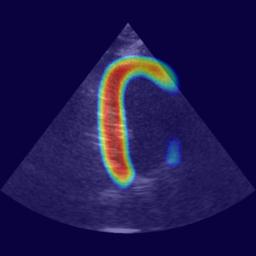}  &     
\includegraphics[trim=40 30 30 40, clip, width=0.1\textwidth]{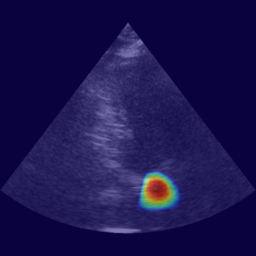}   \\   

\raisebox{0.8\height}{\rotatebox[origin=c]{90}{\scriptsize{MISSForm.}}} & 
\includegraphics[trim=50 50 50 50, clip, width=0.1\textwidth]{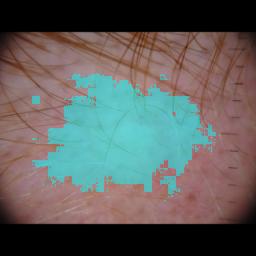} &     
\includegraphics[trim=50 50 50 50, clip, width=0.1\textwidth]{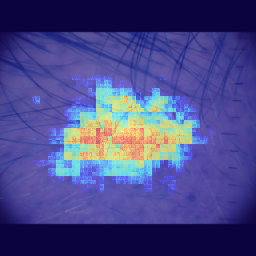} &     
\includegraphics[trim=40 36 25 29, clip,width=0.1\textwidth]{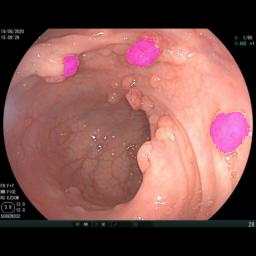} &     
\includegraphics[trim=40 36 25 29, clip,width=0.1\textwidth]{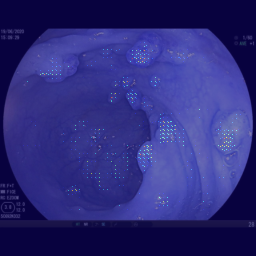} &     
\includegraphics[trim=40 36 25 29, clip,width=0.1\textwidth]{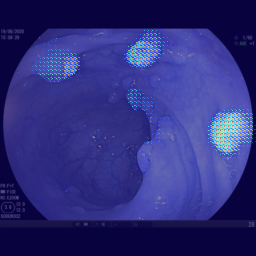} &     
\includegraphics[trim=40 30 30 40, clip, width=0.1\textwidth]{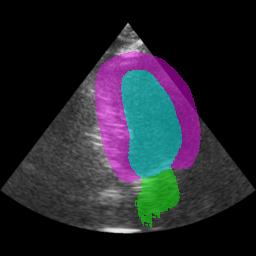} &             
\includegraphics[trim=40 30 30 40, clip, width=0.1\textwidth]{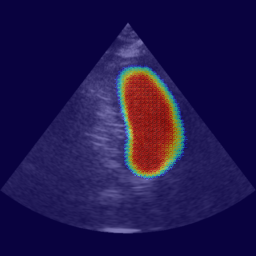} &      
\includegraphics[trim=40 30 30 40, clip, width=0.1\textwidth]{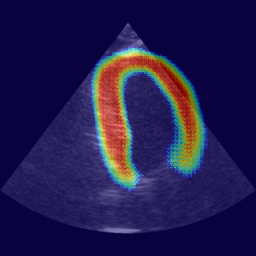}  &     
\includegraphics[trim=40 30 30 40, clip, width=0.1\textwidth]{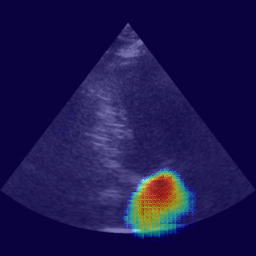}   \\   

\raisebox{0.8\height}{\rotatebox[origin=c]{90}{\scriptsize{\ac{swinumamba}}}} & 
\includegraphics[trim=50 50 50 50, clip, width=0.1\textwidth]{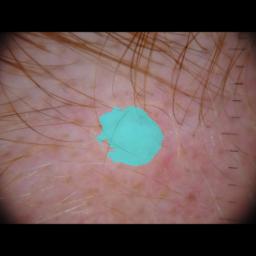} &     
\includegraphics[trim=50 50 50 50, clip, width=0.1\textwidth]{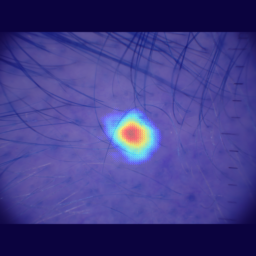} &     
\includegraphics[trim=40 36 25 29, clip, width=0.1\textwidth]{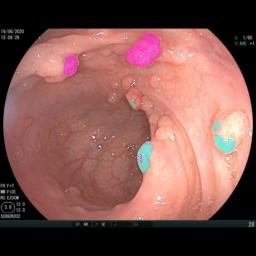} &     
\includegraphics[trim=40 36 25 29, clip, width=0.1\textwidth]{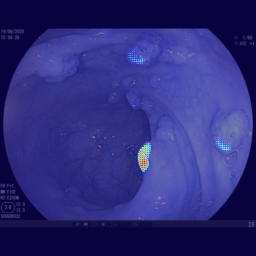} &     
\includegraphics[trim=40 36 25 29, clip, width=0.1\textwidth]{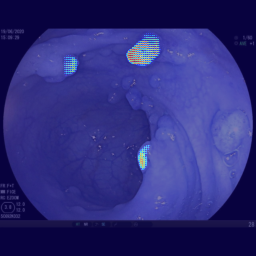} &     
\includegraphics[trim=40 30 30 40, clip, width=0.1\textwidth]{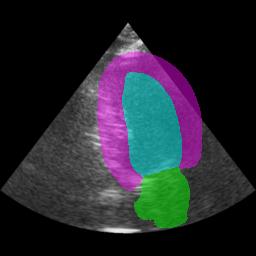} &             
\includegraphics[trim=40 30 30 40, clip, width=0.1\textwidth]{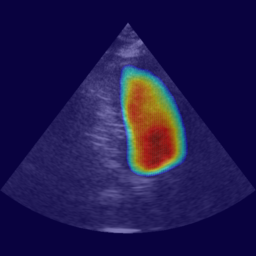} &      
\includegraphics[trim=40 30 30 40, clip, width=0.1\textwidth]{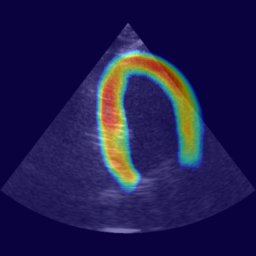}  &     
\includegraphics[trim=40 30 30 40, clip, width=0.1\textwidth]{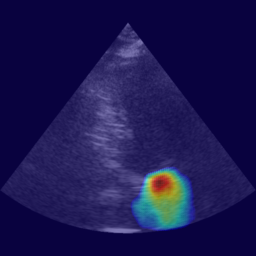}   \\   

\raisebox{1.15\height}{\rotatebox[origin=c]{90}{\scriptsize{\ac{cenet}}}} & 
\includegraphics[trim=50 50 50 50, clip, width=0.1\textwidth]{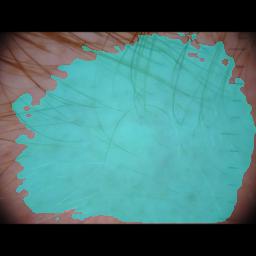} &     
\includegraphics[trim=50 50 50 50, clip, width=0.1\textwidth]{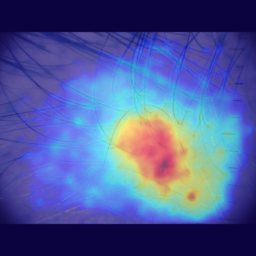} &     
\includegraphics[trim=40 36 25 29, clip, width=0.1\textwidth]{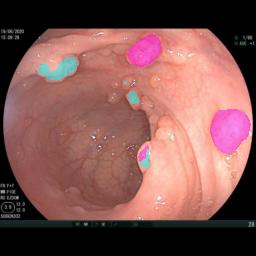} &     
\includegraphics[trim=40 36 25 29, clip, width=0.1\textwidth]{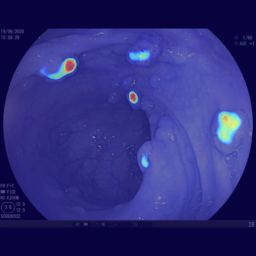} &     
\includegraphics[trim=40 36 25 29, clip, width=0.1\textwidth]{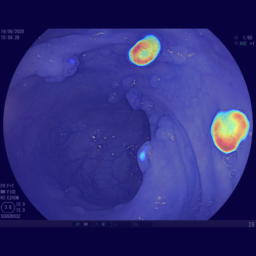} &     
\includegraphics[trim=40 30 30 40, clip, width=0.1\textwidth]{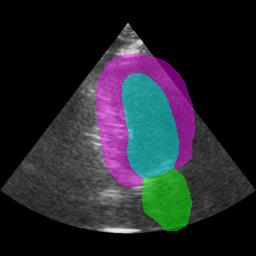} &             
\includegraphics[trim=40 30 30 40, clip, width=0.1\textwidth]{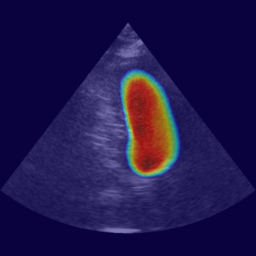} &      
\includegraphics[trim=40 30 30 40, clip, width=0.1\textwidth]{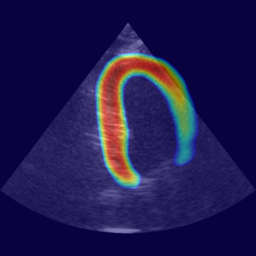}  &     
\includegraphics[trim=40 30 30 40, clip, width=0.1\textwidth]{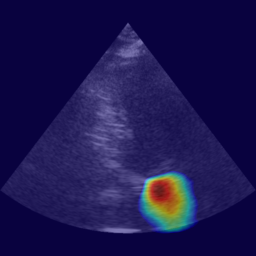}   \\   

\raisebox{0.9\height}{\rotatebox[origin=c]{90}{\scriptsize{\ac{segformer}}}} & 
\includegraphics[trim=50 50 50 50, clip, width=0.1\textwidth]{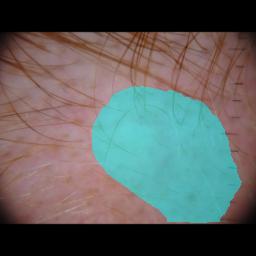} &     
\includegraphics[trim=50 50 50 50, clip, width=0.1\textwidth]{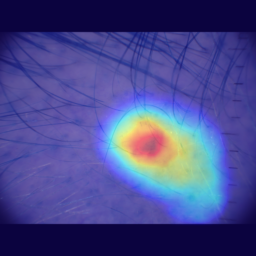} &     
\includegraphics[trim=40 36 25 29, clip, width=0.1\textwidth]{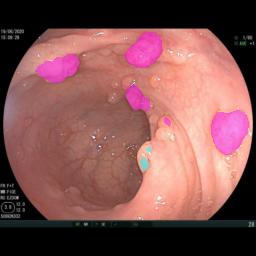} &     
\includegraphics[trim=40 36 25 29, clip, width=0.1\textwidth]{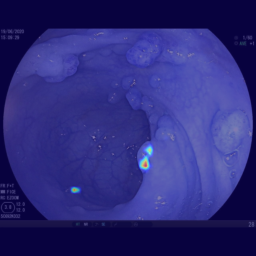} &     
\includegraphics[trim=40 36 25 29, clip, width=0.1\textwidth]{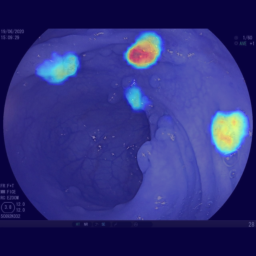} &     
\includegraphics[trim=40 30 30 40, clip, width=0.1\textwidth]{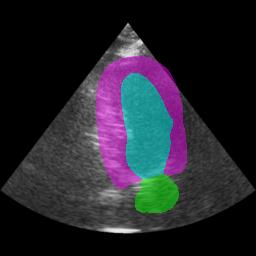} &             
\includegraphics[trim=40 30 30 40, clip, width=0.1\textwidth]{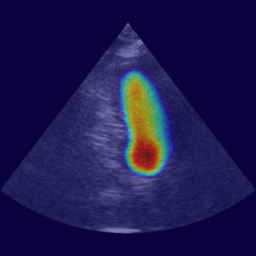} &      
\includegraphics[trim=40 30 30 40, clip, width=0.1\textwidth]{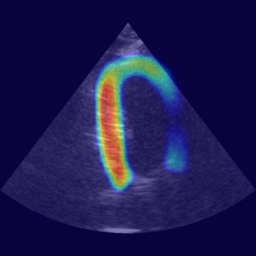}  &     
\includegraphics[trim=40 30 30 40, clip, width=0.1\textwidth]{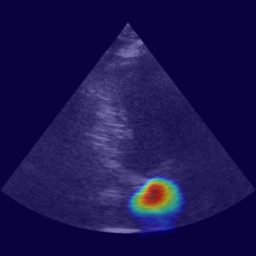}   \\   

\raisebox{1\height}{\rotatebox[origin=c]{90}{\scriptsize{\ac{segnext}}}} & 
\includegraphics[trim=50 50 50 50, clip, width=0.1\textwidth]{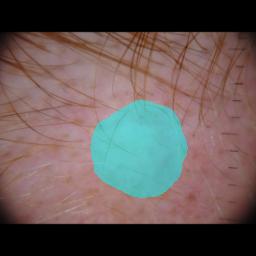} &     
\includegraphics[trim=50 50 50 50, clip, width=0.1\textwidth]{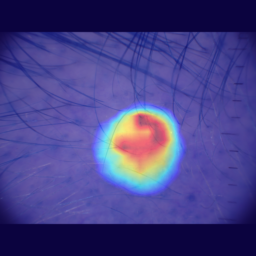} &     
\includegraphics[trim=40 36 25 29, clip, width=0.1\textwidth]{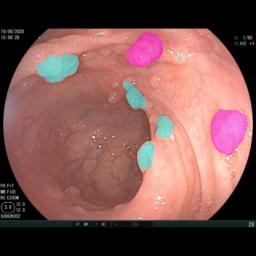} &     
\includegraphics[trim=40 36 25 29, clip, width=0.1\textwidth]{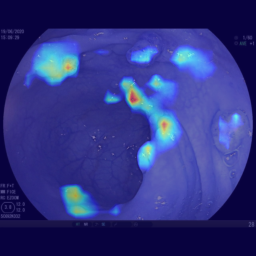} &     
\includegraphics[trim=40 36 25 29, clip, width=0.1\textwidth]{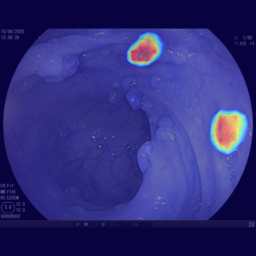} &     
\includegraphics[trim=40 30 30 40, clip, width=0.1\textwidth]{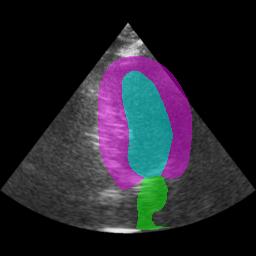} &             
\includegraphics[trim=40 30 30 40, clip, width=0.1\textwidth]{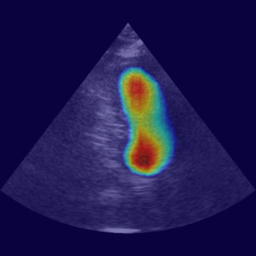} &      
\includegraphics[trim=40 30 30 40, clip, width=0.1\textwidth]{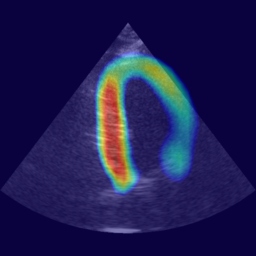}  &     
\includegraphics[trim=40 30 30 40, clip, width=0.1\textwidth]{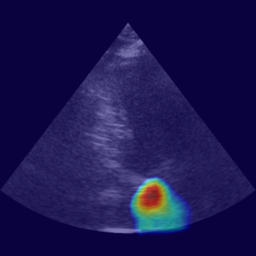}   \\   

\raisebox{1\height}{\rotatebox[origin=c]{90}{\scriptsize{\acs{vwconv}}}} & 
\includegraphics[trim=50 50 50 50, clip, width=0.1\textwidth]{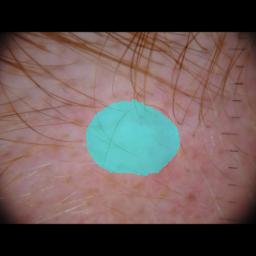} &     
\includegraphics[trim=50 50 50 50, clip, width=0.1\textwidth]{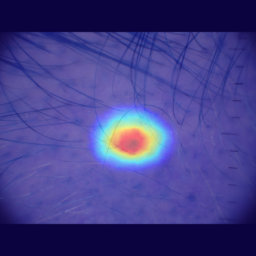} &     
\includegraphics[trim=40 36 25 29, clip, width=0.1\textwidth]{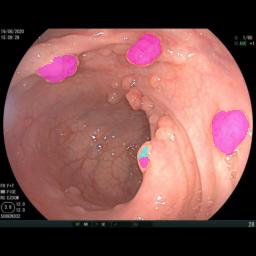} &     
\includegraphics[trim=40 36 25 29, clip, width=0.1\textwidth]{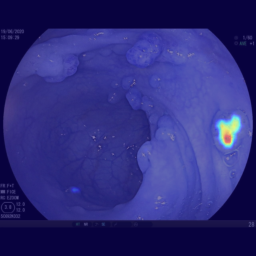} &     
\includegraphics[trim=40 36 25 29, clip, width=0.1\textwidth]{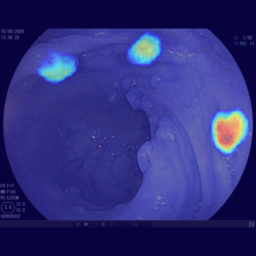} &     
\includegraphics[trim=40 30 30 40, clip, width=0.1\textwidth]{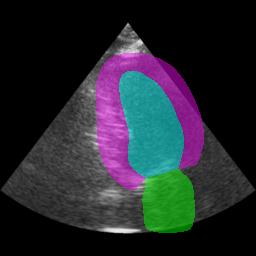} &             
\includegraphics[trim=40 30 30 40, clip, width=0.1\textwidth]{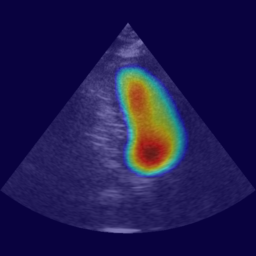} &      
\includegraphics[trim=40 30 30 40, clip, width=0.1\textwidth]{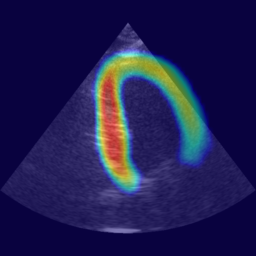}  &     
\includegraphics[trim=40 30 30 40, clip, width=0.1\textwidth]{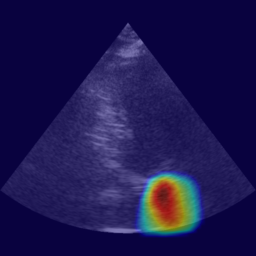}   \\   

\raisebox{1\height}{\rotatebox[origin=c]{90}{\scriptsize{\acs{vwmit}}}} & 
\includegraphics[trim=50 50 50 50, clip, width=0.1\textwidth]{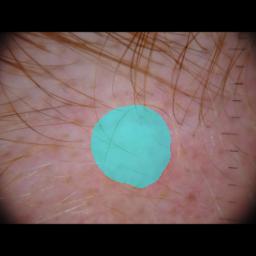} &     
\includegraphics[trim=50 50 50 50, clip, width=0.1\textwidth]{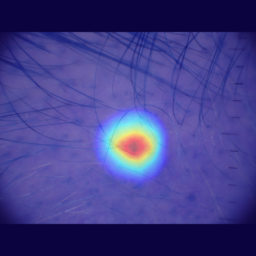} &     
\includegraphics[trim=40 36 25 29, clip, width=0.1\textwidth]{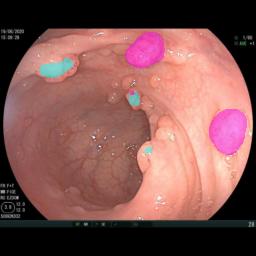} &     
\includegraphics[trim=40 36 25 29, clip, width=0.1\textwidth]{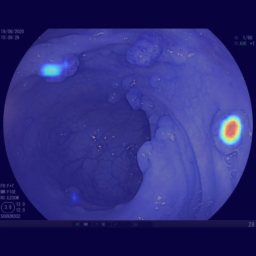} &     
\includegraphics[trim=40 36 25 29, clip, width=0.1\textwidth]{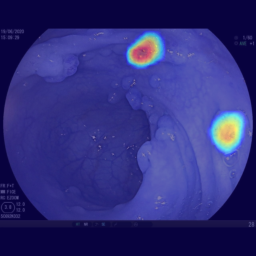} &     
\includegraphics[trim=40 30 30 40, clip, width=0.1\textwidth]{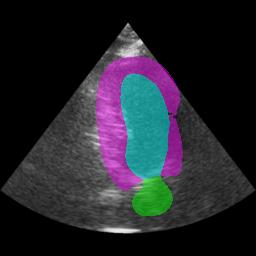} &             
\includegraphics[trim=40 30 30 40, clip, width=0.1\textwidth]{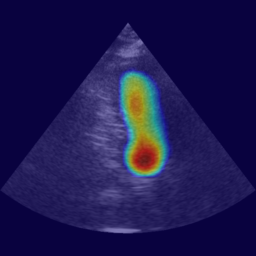} &      
\includegraphics[trim=40 30 30 40, clip, width=0.1\textwidth]{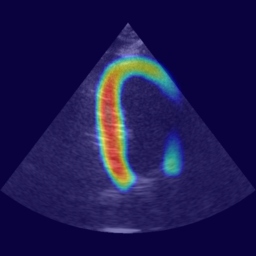}  &     
\includegraphics[trim=40 30 30 40, clip, width=0.1\textwidth]{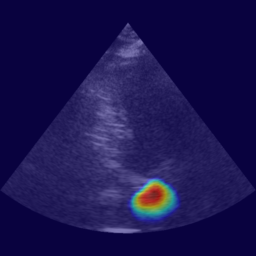}   \\   

\raisebox{0.8\height}{\rotatebox[origin=c]{90}{\scriptsize{InternIm.}}} & 
\includegraphics[trim=50 50 50 50, clip, width=0.1\textwidth]{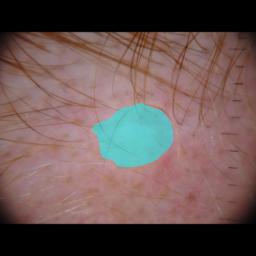} &     
\includegraphics[trim=50 50 50 50, clip, width=0.1\textwidth]{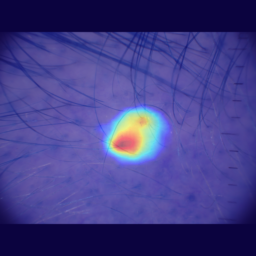} &     
\includegraphics[trim=40 36 25 29, clip, width=0.1\textwidth]{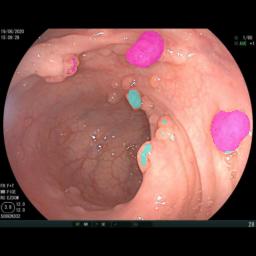} &     
\includegraphics[trim=40 36 25 29, clip, width=0.1\textwidth]{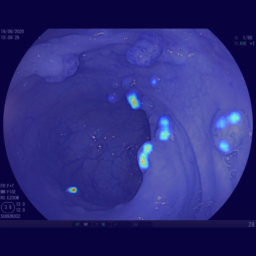} &     
\includegraphics[trim=40 36 25 29, clip, width=0.1\textwidth]{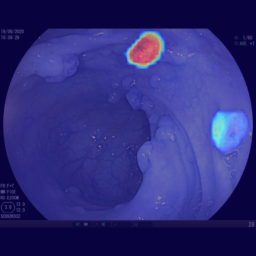} &     
\includegraphics[trim=40 30 30 40, clip, width=0.1\textwidth]{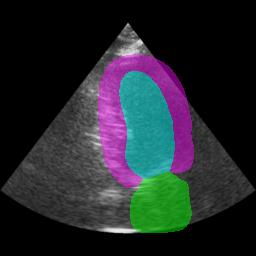} &             
\includegraphics[trim=40 30 30 40, clip, width=0.1\textwidth]{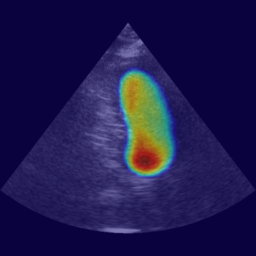} &      
\includegraphics[trim=40 30 30 40, clip, width=0.1\textwidth]{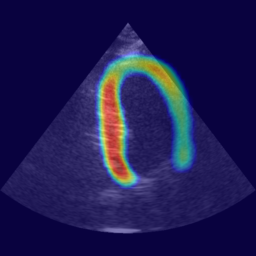}  &     
\includegraphics[trim=40 30 30 40, clip, width=0.1\textwidth]{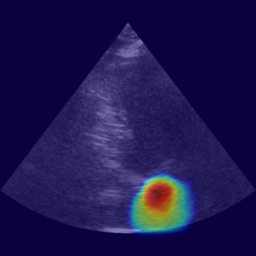}   \\   

\raisebox{0.7\height}{\rotatebox[origin=c]{90}{\scriptsize{\ac{transnext}}}} & 
\includegraphics[trim=50 50 50 50, clip, width=0.1\textwidth]{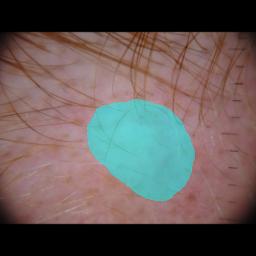} &     
\includegraphics[trim=50 50 50 50, clip, width=0.1\textwidth]{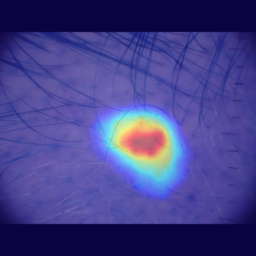} &     
\includegraphics[trim=40 36 25 29, clip, width=0.1\textwidth]{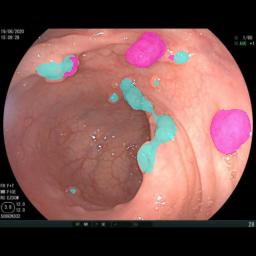} &     
\includegraphics[trim=40 36 25 29, clip, width=0.1\textwidth]{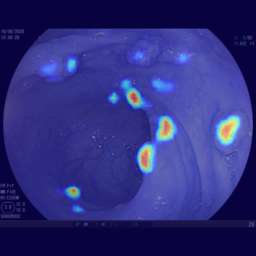} &     
\includegraphics[trim=40 36 25 29, clip, width=0.1\textwidth]{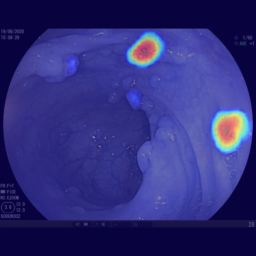} &     
\includegraphics[trim=40 30 30 40, clip, width=0.1\textwidth]{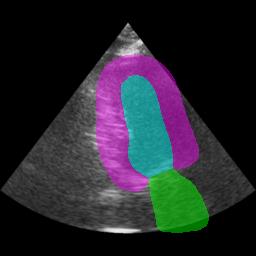} &             
\includegraphics[trim=40 30 30 40, clip, width=0.1\textwidth]{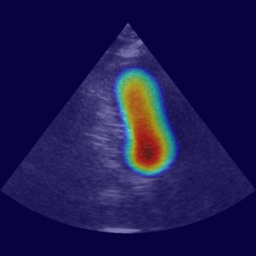} &      
\includegraphics[trim=40 30 30 40, clip, width=0.1\textwidth]{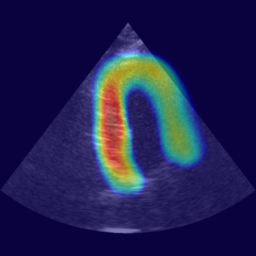}  &     
\includegraphics[trim=40 30 30 40, clip, width=0.1\textwidth]{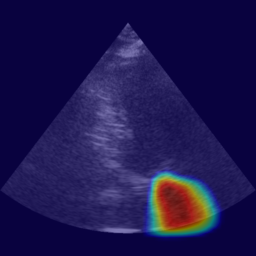}   \\   

\end{tabular}
\end{adjustbox}
\caption{Grad-CAM comparison across datasets and architectures.}
\label{fig:eval:grad_cam:mixed}
\end{figure*}

Figure~\ref{fig:eval:grad_cam:mixed} presents ground truth (GT), model predictions, and corresponding \ac{gradcam} maps for each class, selected from the 50 worst-performing cases per fold that were consistently challenging across all architectures. In the illustrated cases, \acp{gpvm} often exhibited class-discriminative activation in clinically relevant regions despite lacking a dedicated \ac{mis}-specific design. In some cases, their activations appeared more spatially focused than those of certain \acp{sma}, as illustrated by the \ac{isic} example. 
On \ac{bkai}, all methods struggled with non-neoplastic polyps ($C_1$, \textit{cyan}), both in segmentation output and in the corresponding heatmaps, which is consistent with earlier findings. \ac{missformer}, along with several other architectures such as \acs{vwconv} and \acs{vwmit}, showed little to no activation for this class or had activation concentrated in incorrect spatial regions (i.e., $C_2$, \textit{magenta}). 
In the \ac{camus} example, the LV wall ($C_2$) was particularly challenging for \acs{hiformerRN}, \ac{segformer}, and \ac{vwmit}, as reflected by weak or spatially incomplete class-discriminative activation. Conversely, \mbox{\ac{unet}} failed to identify the left atrium ($C_3$, \textit{green}) altogether.
Overall, no consistent superiority of either model family was evident across the qualitatively assessed cases.

\section{Discussion}
\label{sec:discussion}

\textbf{Main findings.} 
This study investigated how modern \acp{gpvm} compare to domain-specific \acp{sma} in 2D \ac{mis} under strictly standardized training and evaluation. Under the controlled in-distribution setting, \acp{gpvm} and \acp{sma} achieved broadly comparable segmentation performance, with no compelling evidence of a practically meaningful model-family advantage. 
While the descriptive cross-dataset summaries showed slightly higher numerical pooled DSC point estimates for several \acp{gpvm}, the adjusted statistical analysis estimated a very small effect in the opposite direction, favoring \acp{sma}, even though this cannot be considered robust evidence due to the unsupported maximal random-effects structure. Overall, both perspectives point to the same broader conclusion: Differences between model families were small and unlikely to represent a practically relevant performance gap in the present benchmark.

\textbf{Interpretation of results.}
The apparent divergence between the descriptive summaries and the LMM results is informative rather than contradictory. The descriptive analysis identifies which models achieve the highest pooled DSC estimates at the dataset level and across classes. In contrast, the LMM evaluates whether a systematic effect of model family persists after accounting for variability attributable to dataset, \ac{cv} fold, and model-level variability. The resulting coefficient for \textit{ModelFamily} was statistically significant but very small ($\beta = 0.006$, $d = 0.04$), suggesting negligible practical relevance. This interpretation is further supported by the similar BIC values of the null and fixed-effects models ($\Delta BIC < 2$), indicating that the inclusion of \textit{ModelFamily} contributed little additional explanatory power beyond the variance already accounted for by the random effects. In addition, the random-effects structure had to be simplified because the data did not support a maximal specification, which may increase the risk of anti-conservative inference due to p-value inflation (Type I error)~\cite{barr2013}. Consequently, the statistical result should not be interpreted as robust evidence that \acp{sma} generally outperform \acp{gpvm}; rather, it reinforces the conclusion that any family-level difference observed here is weak.

This interpretation is consistent with broader concerns regarding evaluation reliability in \ac{mis} and machine learning more broadly. Prior work has shown that model performance can be strongly influenced by methodological choices, including preprocessing, augmentation, hyperparameter tuning, and random initialization~\cite{renard2020variability, varoquaux2022machine, pineau2021improving}. Moreover, recent analyses of \ac{mis} literature indicate that small differences in reported DSC values are often interpreted without adequate consideration of performance variability or statistical uncertainty~\cite{christodoulou2024confidence}. 
Our results reinforce these concerns: Apparent differences based on point estimates were small, dataset-dependent, and not consistently reflected across the adjusted model-family analysis. This supports the view that marginal architectural gains should be interpreted cautiously unless they are robust under controlled and statistically informed evaluation. 

In line with this perspective, the absence of a clear performance advantage for either model family is also consistent with recent evidence that generalist models can achieve competitive performance in \ac{mis}~\cite{moglia2026generalistModelsInMIS}. However, unlike survey-based comparisons that often rely on results obtained under heterogeneous protocols, our study evaluated both model families within a unified pipeline, providing a more controlled basis for comparison.

\textbf{Implications for model development.}
Against this background, our findings call into question the practice of proposing increasingly tailored \acp{sma} without systematically evaluating them against both strong \acp{gpvm} and established \acp{sma} and without adequately accounting for statistical uncertainty. 
Importantly, our study does not imply that \acp{sma} are unnecessary. Medical imaging tasks differ substantially in modality, dimensionality, target size, annotation quality, class imbalance, and clinical constraints. In such settings, task-specific inductive biases may still be valuable. Our results suggest, however, that their added value should be demonstrated empirically under rigorous benchmarking conditions, including strong baselines, appropriate variability estimates, and, where applicable, statistical testing, rather than inferred from architectural novelty or small improvements in mean, median, or point-estimate performance alone.

As discussed above, a strict evaluation protocol, including statistics, is central when assessing whether an apparent model improvement reflects genuine architectural advantage.
The practical implication is a more resource-conscious view of model development in 2D \ac{mis}. 
Before proposing a new \ac{sma}, strong \acp{gpvm} and existing \acp{sma} should be tested to determine whether they are viable candidates for the intended use case. If they achieve comparable performance, research effort may be better directed toward factors with potentially larger clinical impact, such as data quality, annotation consistency, robust training protocols, uncertainty estimation, calibration, interpretability, and \ac{ood} generalization. In this sense, our results should be understood as a thought-provoking benchmark in the context of the evaluated settings rather than a definitive ranking of model families.

\textbf{Threats to validity.}
Our findings are limited to the selected datasets, architectures, and standardized protocol. They reflect in-distribution \ac{cv} and bootstrap resampling and therefore do not assess cross-domain \ac{ood} generalization.
In addition, the benchmark deliberately prioritized experimental depth (i.e., learning-rate tuning for all models and five-fold \ac{cv}) over exhaustive dataset coverage. Although it includes three heterogeneous 2D \ac{mis} datasets, it does not represent the full diversity of clinical imaging. In particular, the results do not support conclusions for volumetric 3D segmentation (e.g., CT, MRI, or PET), extreme low-annotation regimes, pathological subtyping in general, rare anatomical and pathological structures, or strong domain shifts. 
Furthermore, no unified protocol can provide a perfectly fair comparison: Reproducing each model's original training recipe would conflate architecture with recipe-specific ``pipeline advantages'', whereas standardization may disadvantage models designed for different settings. Our protocol should therefore be interpreted as a best-effort controlled family-level benchmark, rather than evidence that each model reached its optimal performance.
Similarly, the ImageNet-pretraining condition was standardized across the benchmark. Its potential homogenizing effect cannot be isolated in the present design and would require dedicated comparisons, for example, with training from scratch. 
Accordingly, the possibility that large-scale pretraining attenuates family-level differences cannot be ruled out; if so, evaluating readily available pretrained architectures before developing and training a new \ac{sma} from scratch would be further justified.

The selected model set is necessarily incomplete, and different hyperparameter budgets or architecture-specific training strategies could change the observed results. Furthermore, \textit{ModelFamily} is a coarse grouping that may obscure important differences between individual architectures; consequently, the results do not imply definitive architectural equivalence. Finally, the statistical analysis was constrained by the available data structure and required a simplified random-effects specification, so the detected fixed effect should be interpreted cautiously and must not be read as robust evidence that \acp{sma} outperform \acp{gpvm}.

\textbf{Future work.}
Future studies should extend the benchmark to additional 2D modalities, including X-ray, fundus, and microscopy, as well as to further architectures and input resolutions. A dedicated study is required to examine whether the findings extend to volumetric 3D CT, MRI, or PET segmentation, where explicit 3D priors may be advantageous.
More detailed analyses should also investigate under which conditions one model family may be preferable, for example with respect to target size, class imbalance, annotation noise, pretraining, or domain shift. 
Evaluating \ac{ood} generalization on related held-out datasets, such as Kvasir-SEG alongside \ac{bkai}, is an essential next step. 
The version of the benchmarking tool used for this study is already available at \href{\githuburl}{GitHub}; a curated and documented open-source release will support reproducibility and community-driven extensions~\cite{borst2026medsegBenchmarker}.

\section{Conclusion}

Under standardized training and evaluation conditions, modern \acp{gpvm} and domain-specific \acp{sma} achieved broadly comparable performance in our evaluated 2D \ac{mis} settings. Our results do not provide evidence for a practically meaningful performance difference between the two model families and therefore do not support a clear claim of model-family superiority. Although several \acp{gpvm} achieved slightly higher descriptive pooled DSC summaries, the adjusted statistical analysis indicated at most a very small, non-robust model-family effect in favor of \acp{sma}, with negligible effect size and relevant modeling caveats.
The central message is therefore not that one family should replace the other, but that claims of architectural superiority in \ac{mis} require strong baselines, controlled protocols, variability estimates, and statistical evaluation, rather than relying on marginal improvements in mean, median, or point estimates alone.
While our findings should not be extrapolated to 3D segmentation, extreme low-annotation regimes, pathological subtyping, or \ac{ood} generalization, they encourage a critical reassessment of routine architectural innovation in 2D \ac{mis} and motivate broader analyses before drawing definitive conclusions about model-family advantages.

\section*{Acknowledgements}
The authors would like to thank Dr. Dominik Müller (University of Augsburg) for his valuable and insightful advice during the preparation of this manuscript. In addition, we sincerely thank Timo Dittus for his Master's thesis, upon which our codebase is partially built.
\bibliography{bib_main}


\clearpage
\appendix

\vspace{2em}

\runninghead{Borst et al.}{Benchmarking GP-VMs and SMAs for 2D MIS (Suppl.)}

\makeappendixtitle

\counterwithin{figure}{section} 
\counterwithin{table}{section} 
\renewcommand{\thefigure}{\thesection\arabic{figure}}
\renewcommand{\thetable}{\thesection\arabic{table}}

\section{Overview of the Included Architectures}

Below, we briefly summarize the key design choices of each evaluated architecture and document the implementation details used in our benchmark. The corresponding code and configurations are available at \href{\githuburl}{GitHub}. Unless stated otherwise, we follow the architecture-specific settings recommended by the respective authors or official implementations.

\subsection{\Aclp*[uppercase/cmd=\ecapitalisewords]{sma}}

\paragraph{\acl{unet}~\cite{appendix_ronneberger2015unet}.}
U-Net is a fully convolutional encoder-decoder network with a symmetric, multi-resolution design. Its contracting path successively applies convolutional blocks and max-pooling operations to enlarge the receptive field, whereas its expanding path restores spatial resolution through transposed convolutions followed by convolutional refinement. Feature maps from matching encoder levels are concatenated with decoder features through skip connections, which constitute the model's central mechanism for preserving fine spatial information across the decoding pathway. 
\textit{For a comparable initialization condition, the convolutional layers of the first four encoder stages were initialized with ImageNet-pretrained VGG13 weights, while the bottleneck, decoder, and head were trained from scratch. Notably, we used same-padded $3\times3$ convolutions (\texttt{padding=1}) to preserve spatial dimensions and avoid cropping encoder features before skip concatenation.}

\paragraph{\acl{hiformerRN}~\cite{appendix_heidari2023hiformer}.}
HiFormer combines a CNN feature pyramid with a coupled hierarchical Swin Transformer pathway. At each scale, a $1\times1$ projection transfers CNN features to the corresponding transformer level, where their element-wise addition injects local spatial information into the transformer representation.
Subsequent patch-merging stages propagate the fused representation toward coarser scales.
A double-level fusion (DLF) module then exchanges information between the finest- and coarsest-resolution representations by cross-attention. The resulting features are decoded using a lightweight convolutional upsampling path. Thus, HiFormer explicitly combines convolutional locality with windowed self-attention and cross-scale fusion. 
\textit{We used the HiFormer-B configuration, but replaced its default ResNet50 encoder with ResNet101 (denoted as \acs{hiformerRN}) to obtain a capacity closer to that of the other models. The ResNet101 backbone was truncated after its third residual stage; consequently, its \(1/4\)-, \(1/8\)-, and \(1/16\)-resolution feature maps were used, while the final \(1/32\)-resolution stage was omitted. 
Although the public repository defaults to \texttt{num\_heads=(6,12)}, we used \texttt{num\_heads=(6,6)} and gave the paper precedence because it explicitly states: ``We observe that the pair of (2, 1) for (S, L) and six heads for both levels work best.'' Since the public implementation assumed $224\times224$ inputs, we adapted its resolution-dependent components to support $256\times256$.}

\paragraph{\acl{missformer}~\cite{appendix_huang2022missformer}.}
\ac{missformer} is a hierarchical transformer-based segmentation architecture comprising a Mix Transformer (MiT) encoder, a multi-scale feature bridge, and a transformer-based decoder. 
Following an initial overlapping patch embedding, successive transformer stages and overlapping patch-merging operations construct a multi-scale encoder pyramid. 
Each transformer block combines efficient self-attention with an Enhanced Mix-FFN (ReMix-FFN), which reintroduces local spatial information through depth-wise convolution while complementing the global interactions captured by self-attention.
Between encoder and decoder, the ReMixed Transformer Context Bridge jointly processes features from all scales, thereby modeling local, global, and cross-scale correlations. 
The decoder progressively reconstructs the prediction resolution using patch-expansion layers and transformer blocks with ReMix-FFN. Finally, a linear projection produces per-pixel segmentation logits.
\textit{Unlike the original paper, which trained \ac{missformer} from scratch, we initialized the MiT-B1 encoder with ImageNet-pretrained weights. Since the public implementation of \ac{missformer} contained fixed-resolution assumptions, we adapted these settings to the common $256\times256$ benchmark input described above.}

\paragraph{\acl{swinumamba}~\cite{appendix_liuSwinUMambaMICCAI2024}.}
\acl{swinumamba} is a U-shaped segmentation method that combines Mamba~\cite{appendix_gu2024mamba}, a selective state-space model, with ImageNet-pretrained visual representations. It comprises a hierarchical Visual State Space (VSS) encoder, a convolutional decoder, and a $1\times1$ convolutional segmentation head.
Each VSS block in the encoder employs a two-dimensional selective state-space module (SS2D) to perform Mamba-style selective scans. Feature maps are traversed in four directions, processed by state-space models, and merged back into a two-dimensional representation. 
This design captures long-range dependencies with linear rather than quadratic sequence complexity.
The encoder produces a five-stage feature hierarchy comprising a convolutional stem and four VSS stages. 
A $2\times2$ patch embedding and subsequent patch-merging layers progressively reduce spatial resolution while increasing the feature dimension.
Overall, the encoder structure closely follows VMamba-Tiny~\cite{appendix_liu2024vmamba}, enabling initialization of compatible VSS blocks and patch-merging layers from ImageNet-1K-pretrained weights.
The decoder fuses encoder features through corresponding-scale skip connections, refines them using residual convolutional blocks, and progressively upsamples them with transposed convolutions.
\textit{We used the standard configuration rather than the lighter Mamba-decoder variant, Swin-UMamba$^{\dagger}$. Compatible VSS-block and patch-merging weights were initialized from the ImageNet-1K-pretrained VMamba-Tiny checkpoint. Although the original Swin-UMamba paper describes auxiliary segmentation heads for deep supervision, we retained the reference implementation's default setting, which is \texttt{deep\_supervision=False}, disabling deep supervision.}

\paragraph{\acl{cenet}~\cite{appendix_bozorgpour2025CENet}.}
The Context Enhancement Network (CENet) combines a hierarchical Pyramid Vision Transformer V2 (PVTv2) encoder for multi-scale feature extraction with skip connections refined by Dual Selective Enhancement Blocks (DSEBs) to preserve boundary information. Along the decoder pathway, Context Feature Attention Modules (CFAMs) integrate multi-scale contextual information while mitigating feature redundancy and excessive enhancement.
More specifically, each DSEB concatenates encoder and decoder features and applies feature-edge amplification and differential attention to emphasize boundaries and fine details, suppress less relevant contextual information, and highlight salient structures.
At each decoder stage, a CFAM recalibrates channel responses, aggregates multi-scale contextual information using multiple receptive fields, and applies a weighted Non-local Block (wNLB) to reduce noise accumulation. An enhanced \ac{mlp} block equipped with a Spatial Calibration Module (SRM) subsequently refines the feature representations. Overall, these components aim to mitigate information loss at object boundaries and in small structures.
\textit{We instantiated \acs{cenet} with an ImageNet-pretrained PVTv2-B3 backbone rather than the original PVTv2-B2 configuration. Moreover, we generalized the resolution-dependent decoder settings to support additional input resolutions, including the $256\times256$ inputs used in our experiments.
The employed FEA scale factors \texttt{[0.8, 0.4]} and the stage-specific MCA dilation-rate schedule follow the original CENet code repository~\href{https://github.com/xmindflow/cenet/tree/main/src/networks/cenet}{(GitHub)} and differed from the settings reported in the CENet paper.}

\subsection{General-Purpose Vision Models: \Acl*[uppercase/cmd=\ecapitalisewords]{ss}}

\paragraph{\acl{segformer}~\cite{appendix_xie2021segformer}.}
\ac{segformer}'s key design is a hierarchical Mix Transformer (MiT) encoder without positional embeddings, followed by a lightweight All-\ac{mlp} decoder.
The MiT encoder constructs a four-level feature pyramid at $1/4$, $1/8$, $1/16$, and $1/32$ of the input resolution. It uses an initial overlapping patch-embedding layer followed by overlapping patch-merging layers at subsequent encoder-stage transitions.
Each transformer block combines efficient spatial-reduction self-attention, which lowers computational complexity at higher-resolution stages by shortening the sequence length, with a Mix-FFN.
The latter inserts a $3\times3$ depthwise convolution between two MLP layers to introduce local spatial information.
MiT does not employ positional embeddings, thereby avoiding positional-embedding interpolation when test images have a different resolution.
The All-\ac{mlp} decoder linearly projects the multi-scale features from all four encoder stages to a common channel dimension, upsamples them to $1/4$ of the input resolution, and concatenates them. The concatenated features are fused and projected to class-specific logits by pointwise $1\times1$ convolutions, which implement the MLP fusion and prediction layers. Finally, the logits at one-quarter of the input resolution are bilinearly upsampled to the original input resolution.
\textit{We used the SegFormer-B3 variant, whose MiT-B3 encoder has feature dimensions of 64, 128, 320, and 512, with 3, 4, 18, and 3 transformer blocks across its four stages. We initialized the encoder from ImageNet-1K-pretrained weights.}

\paragraph{\acl{segnext}~\cite{appendix_guo2022segnext}.}
\ac{segnext} is a convolutional encoder-decoder architecture comprising a hierarchical Multi-Scale Convolutional Attention Network (MSCAN) encoder and a lightweight Hamburger decoder for global-context modeling.
The MSCAN encoder constructs a four-stage feature pyramid at $1/4$, $1/8$, $1/16$, and $1/32$ of the input resolution. Each stage begins with spatial downsampling and is followed by a stack of convolutional building blocks that are similar to the structure of ViT but replace self-attention with multi-scale convolutional attention (MSCA) and use batch normalization.
Within each block, an MSCA module aggregates local information with a depth-wise convolution, then captures several receptive fields with parallel depth-wise strip convolutions, and finally applies a $1\times1$ convolution to model channel relationships and generate attention weights that are in turn used to re-weight the input of MSCA.
A lightweight Hamburger~\cite{appendix_geng2021hamburger} decoder aggregates the feature maps from the last three encoder stages. Its Hamburger module approximates global context through low-rank matrix decomposition. 
The resulting representation is projected to class-specific logits by $1\times1$ convolutions and bilinearly upsampled to the original input resolution.
\textit{We used the SegNeXt-L variant, whose MSCAN encoder has feature dimensions of 64, 128, 320, and 512, with 3, 5, 27, and 3 convolutional blocks across its four stages. The encoder was initialized with ImageNet-1K-pretrained MSCAN-L weights. The Hamburger decoder used a channel dimension of 1024.}

\paragraph{\acl{vwformer}~\cite{appendix_yan2024vwformer}.}
\acs{vwformer} is a multi-scale decoder designed to improve multi-scale representations produced by hierarchical vision backbones. Varying window attention (VWA) forms its core and provides representations with different receptive-field sizes.
First, \acs{vwformer} aggregates the final three backbone stages using \ac{mlp} projections. More precisely, the features from the $1/16$- and $1/32$-resolution stages are upsampled to the $1/8$ resolution, concatenated with the $1/8$-resolution features, and fused by a further projection to reduce the channel dimensionality.
The resulting representation is processed by three parallel VWA branches with context-window ratios of $2$, $4$, and $8$, complemented by a shortcut path for the most local scale.
VWA keeps queries in local windows but draws keys and values from context windows of different, potentially larger spatial extent, thereby increasing the receptive field. 
To avoid the substantial computational and memory overhead of enlarged context windows, VWA employs the pre-scaling principle and densely overlapping patch embedding (DOPE). In addition, copy-shift padding (CSP) prevents the edge-related attention collapse caused by zero padding.
Finally, the outputs of the VWA branches and the shortcut path are concatenated and projected. 
The resulting representation is then enhanced with low-level features from the first backbone stage using further \ac{mlp} layers. A $1\times1$ convolution produces the class-specific logits, which are subsequently bilinearly upsampled to the input resolution.
\textit{We evaluated VWFormer with an ImageNet-pretrained MiT-B3 backbone (VW-MiT) and an ImageNet-pretrained ConvNeXt-S backbone (VW-Conv). Because the public repository did not specify the exact MiT configuration, we set \texttt{nheads}=1 and enabled shortcut connections for VW-MiT. The ConvNeXt-S configuration used \texttt{nheads}=16 with shortcut connections enabled. Following the original repository implementation, an adaptive-average-pooled global branch was included in the aggregation of multi-scale representations, in addition to the three VWA branches and the shortcut path.}

\subsection{General-Purpose Vision Models: Modern Vision Backbones}

\paragraph{\acl{internimage}~\cite{appendix_wang2023internimage}.}
\ac{internimage} is a hierarchical CNN-based vision foundation model centered on the third-generation deformable convolution (DCNv3). For \ac{ss}, it is coupled with a UPerHead decoder~\cite{appendix_xiao2018uperhead}, as in the original work.
Following an initial convolutional stem that reduces the input resolution by a factor of four, \ac{internimage} comprises four stages of basic blocks at $1/4$, $1/8$, $1/16$, and $1/32$ of the input resolution, with convolutional downsampling layers between adjacent stages.
The \ac{internimage} blocks follow a ViT-inspired design with layer normalization, feed-forward components, and GELU activations, with DCNv3 serving as the core operator. Rather than sampling from a fixed grid, DCNv3 predicts input-dependent spatial offsets and modulation scales using a separable convolution consisting of a $3\times3$ depth-wise convolution followed by separate linear projections. This enables adaptive spatial aggregation and a content-dependent effective receptive field.
DCNv3 further divides the features into aggregation groups, each with independent sampling offsets and modulation scales for a $3\times3$ sampling kernel. This allows the network to capture information from different representation subspaces at different spatial locations.
For \ac{ss}, the UPerHead decoder aggregates the feature maps from all four \ac{internimage} stages using pyramid pooling and a Feature Pyramid Network (FPN)-style top-down pathway. The resulting fused representation is converted into class-specific logits by a $1\times1$ convolution and bilinearly upsampled to the input resolution.
\textit{We used the InternImage-T variant as the backbone, whose four stages have feature dimensions of 64, 128, 256, and 512, with 4, 4, 18, and 4 InternImage blocks and DCNv3 group counts of 4, 8, 16, and 32, respectively. The encoder was initialized with ImageNet-1K-pretrained weights. The UPerHead decoder used 512 channels and pooling scales $(1,2,3,6)$ to convert the feature pyramid into segmentations.}

\paragraph{\acl{transnext}~\cite{appendix_shi2024transnext}.}
\ac{transnext} is a hierarchical vision-transformer backbone that combines Aggregated Attention (AA) as its token mixer with a Convolutional Gated Linear Unit (GLU) as its channel mixer. For \ac{ss}, a UPerHead~\cite{appendix_xiao2018uperhead} decoder can be attached to the resulting feature hierarchy, as in the original work.
Following an initial overlapping patch-embedding layer that reduces the input resolution by a factor of four, \ac{transnext} comprises four stages operating at $1/4$, $1/8$, $1/16$, and $1/32$ of the input resolution. Adjacent stages are connected by overlapping patch-embedding layers that perform spatial downsampling.
Each stage consists of multiple building blocks. The first three stages employ AA, whereas the final stage uses multi-head self-attention at $1/32$ resolution. 
AA follows a dual-path design that combines sliding-window attention for local information with spatial reduction attention for global context. The former provides each query with fine-grained information from neighboring features, while the latter supplies coarse-grained global context from spatially reduced features. AA further incorporates learnable tokens into each attention head to diversify the resulting affinity matrices.
The second core operation in each building block is a Convolutional GLU feed-forward module, which performs gated channel mixing while using a depth-wise $3\times3$ convolution to incorporate local spatial information from neighboring features.
For \ac{ss}, the UPerHead decoder combines feature maps from all four \ac{transnext} stages using pyramid pooling and an FPN-style top-down pathway. The resulting representation is projected to class-specific logits using a $1\times1$ convolution and bilinearly upsampled to the input resolution.
\textit{We used the TransNeXt-Tiny variant, whose four stages have feature dimensions of 72, 144, 288, and 576, with 2, 2, 15, and 2 blocks, respectively. The backbone was initialized with ImageNet-1K-pretrained weights. The UPerHead decoder used 512 channels and pooling scales $(1,2,3,6)$.}

\clearpage
\section{Dataset Class Imbalance}

\begin{figure}[ht]
    \centering
    \includegraphics[width=0.8\textwidth]{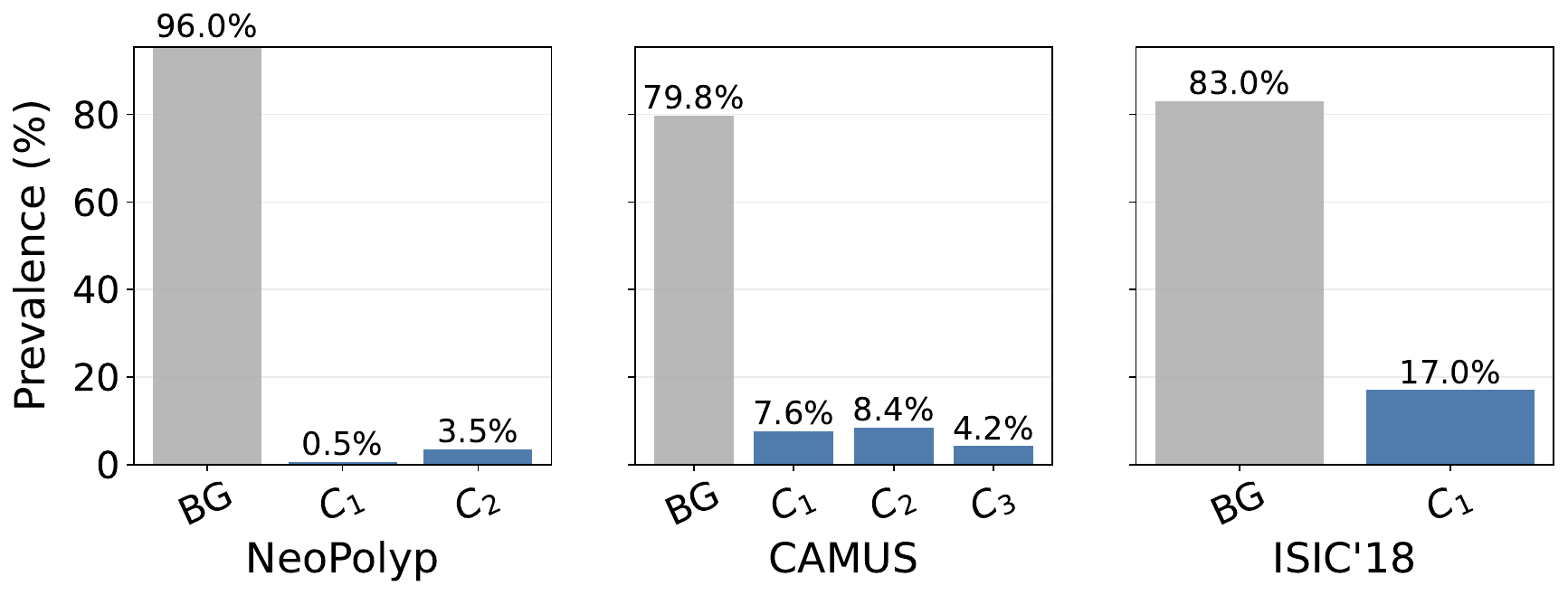}
   \caption{Class-wise GT pixel prevalence across the three datasets (in \%).}
\label{fig:dataset:pixel_count}
\end{figure}

Figure~\ref{fig:dataset:pixel_count} shows the class-wise ground-truth (GT) pixel prevalence across the three datasets: \acs{bkai}, \acs{camus}, and \acs{isic}. All datasets exhibit pronounced class imbalance, with background pixels accounting for the majority of the image area (96.0\% in NeoPolyp, 79.8\% in CAMUS, and 83.0\% in \ac{isic}). In NeoPolyp, the foreground is particularly sparse, with non-neoplastic polyps ($C_1$) and neoplastic polyp tissue ($C_2$) accounting for only 0.5\% and 3.5\% of pixels, respectively. CAMUS exhibits a more moderate imbalance, with the LV wall ($C_2$), LV cavity ($C_1$), and LA ($C_3$) comprising 8.4\%, 7.6\%, and 4.2\% of pixels, respectively. In \ac{isic}, the single foreground class ($C_1$) accounts for 17.0\% of pixels, which still remains underrepresented relative to background (17.0\% vs. 83.0\%). Overall, these distributions demonstrate substantial dataset-dependent variation in class imbalance, which can affect segmentation performance across classes and models. Accordingly, background pixels were excluded from all metric computations to focus the evaluation exclusively on foreground segmentation performance.

\section{Fold-Level Variability}

\begin{figure}[ht]
\begin{center}
    \includegraphics[width=\textwidth]{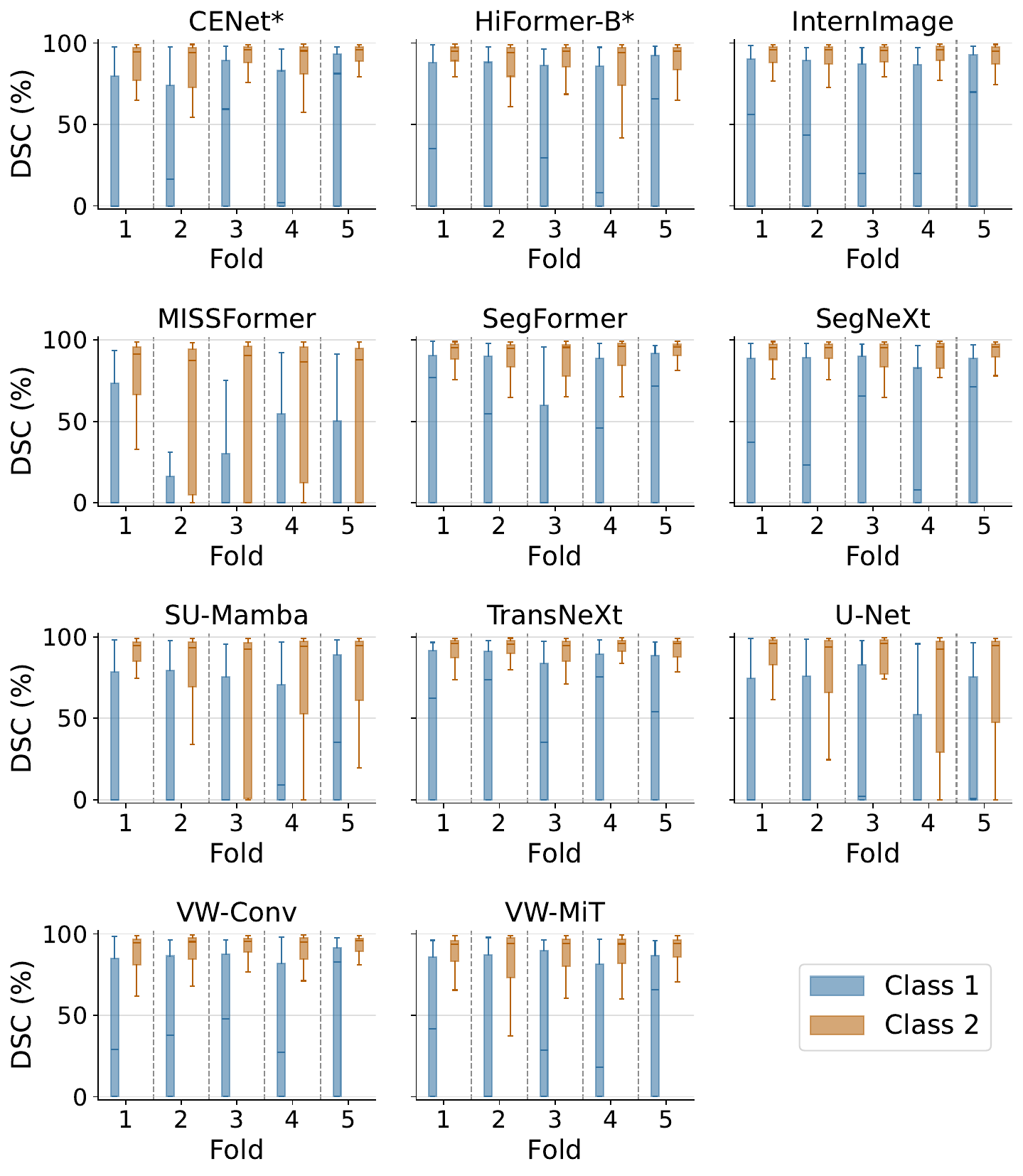}
\end{center}
   \caption{Fold-wise class DSC distribution on the \acs{bkai} dataset.}
\label{fig:boxplot:bkai}
\end{figure}

\begin{figure}[ht]
\begin{center}
    \includegraphics[width=\textwidth]{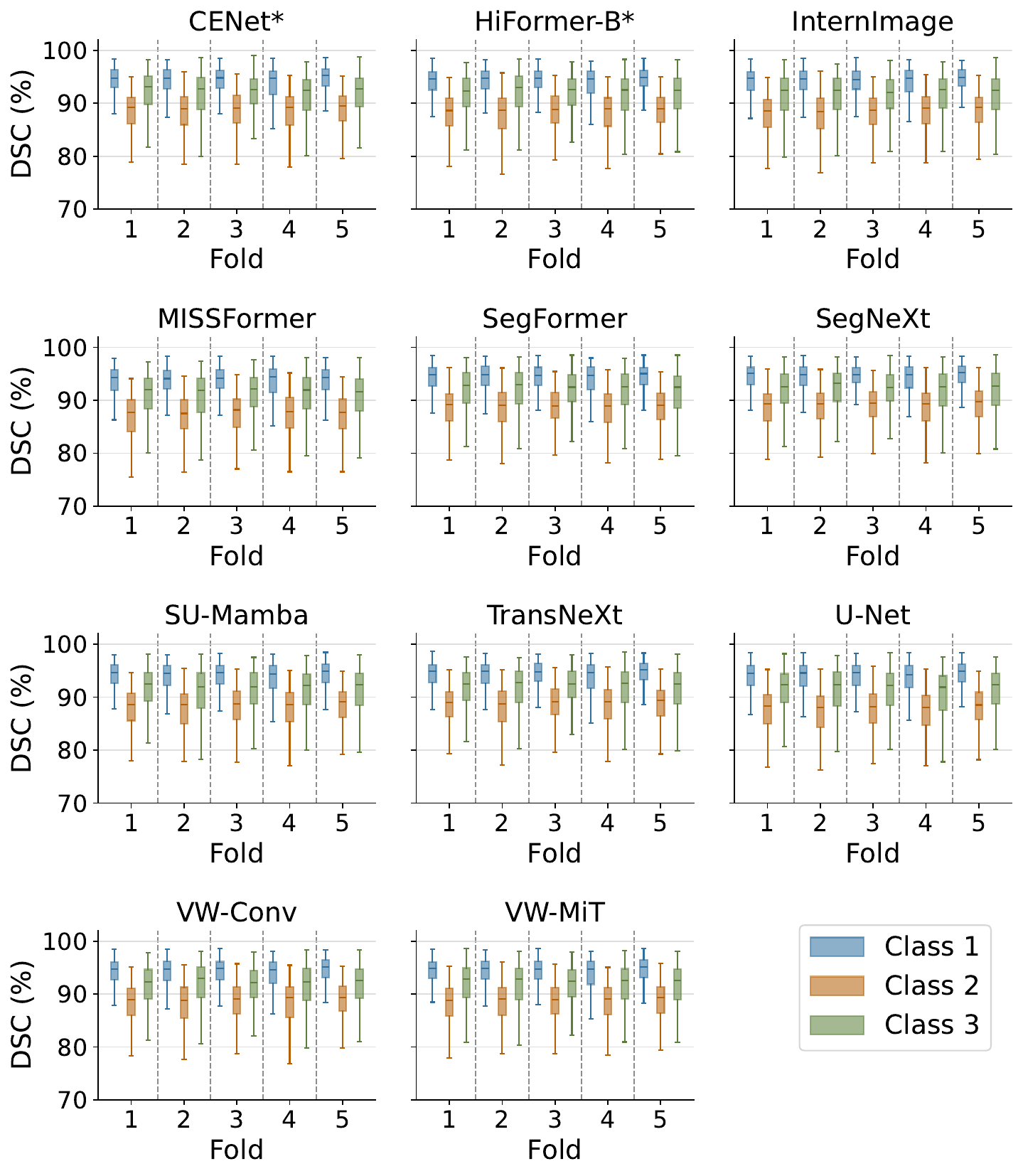}
\end{center}
   \caption{Fold-wise class DSC distribution on the \acs{camus} dataset.}
\label{fig:boxplot:camus}
\end{figure}

\begin{figure}[ht]
\begin{center}
    \includegraphics[width=\textwidth]{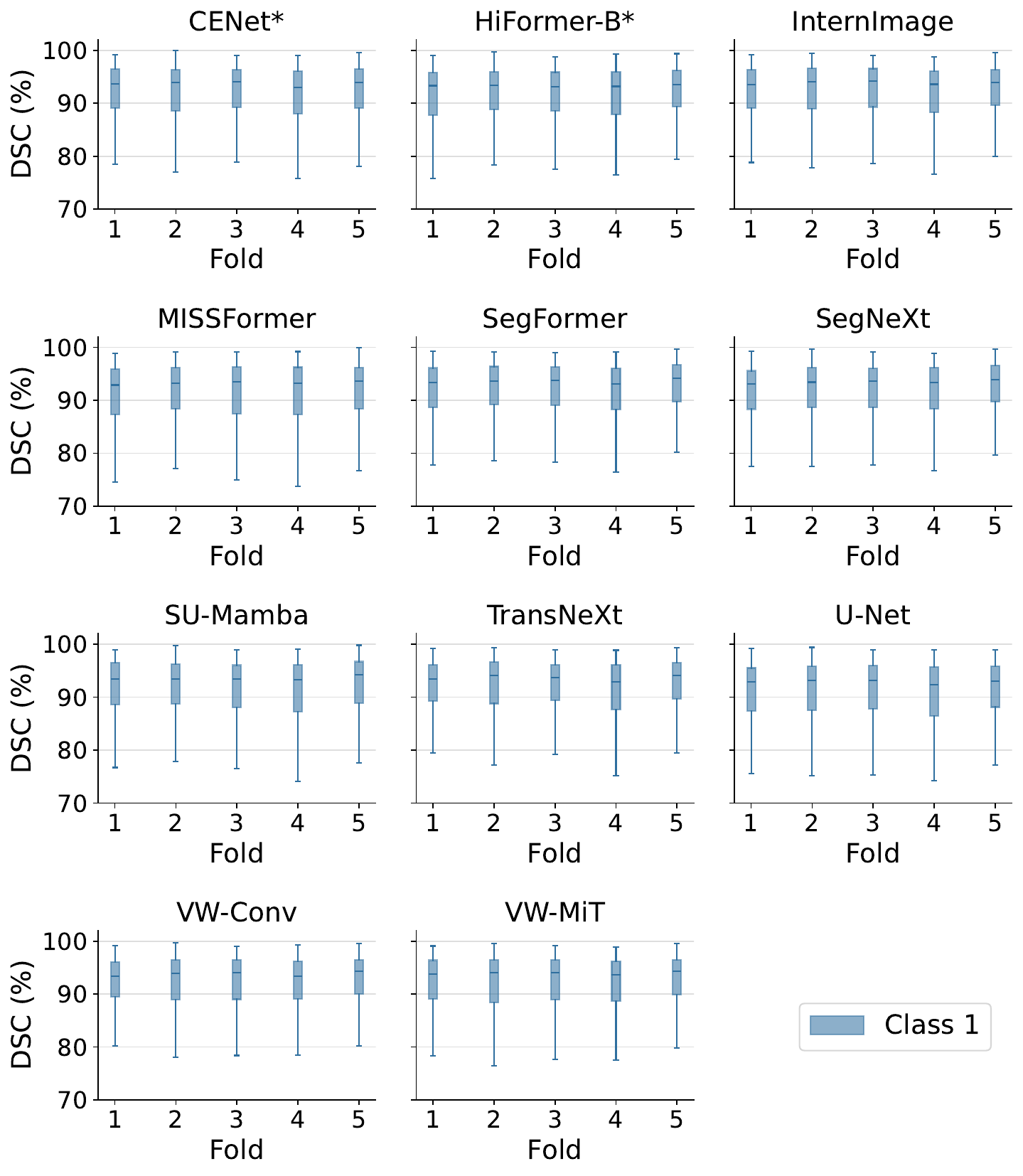}
\end{center}
   \caption{Fold-wise DSC distribution on the \ac{isic} dataset.}
\label{fig:boxplot:isic}
\end{figure}

Figures~\ref{fig:boxplot:bkai} to~\ref{fig:boxplot:isic} depict the fold-wise distributions of image-level foreground class DSC scores for the three datasets. Each subplot represents one architecture, and each fold contains color-coded boxplots for the foreground
classes. DSC was computed per image and class. Cases where a class was absent in both prediction and reference were treated as undefined and excluded. Boxes show the interquartile range with median lines, and whiskers show the non-outlier range, while outliers are omitted for readability. 
For \acs{camus} and \acs{isic}, the y-axis is truncated to 0.7--1.0 for readability because the displayed box-and-whisker ranges consistently lie in this region. 

\end{document}